\documentclass{article}
\usepackage[numbers]{natbib}
\usepackage[final]{neurips_2024}
\usepackage[utf8]{inputenc}
\usepackage[T1]{fontenc}
\usepackage{booktabs}
\usepackage{microtype}
\usepackage{xcolor}
\usepackage{graphicx}
\usepackage{subfig}

\usepackage{algorithmic}
\usepackage{algorithm}

\usepackage{amsmath,amsfonts,bm}

\def\eqref#1{(\ref{#1})}

\def\1{\bm{1}}

\DeclareMathAlphabet{\mathsfit}{\encodingdefault}{\sfdefault}{m}{sl}
\SetMathAlphabet{\mathsfit}{bold}{\encodingdefault}{\sfdefault}{bx}{n}

\def\gM{{\mathcal{M}}}

\def\gX{{\mathcal{X}}}
\def\gY{{\mathcal{Y}}}

\newcommand{\Exs}{\ensuremath{{\mathbb{E}}}}

\newcommand{\numClients}{\ensuremath{M}}

\newcommand{\clientDist}{\ensuremath{\mathcal{P}}}

\newcommand{\modelSize}{m}

\usepackage{amsthm}
\usepackage{amssymb}
\usepackage{mathtools}
\usepackage{wrapfig}
\usepackage{url}
\usepackage{hyperref}
\hypersetup{hidelinks}
\theoremstyle{plain}
\newtheorem{theorem}{Theorem}[section]
\newtheorem{proposition}[theorem]{Proposition}

\theoremstyle{definition}

\newtheorem{assumption}[theorem]{Assumption}
\theoremstyle{remark}

\newcommand{\ie}{\textit{i}.\textit{e}.}
\newcommand{\eg}{\textit{e}.\textit{g}.}

\newcommand{\ours}{\emph{LSS}}

\title{Local Superior Soups: A Catalyst for Model Merging in Cross-Silo Federated Learning}

\author{
  Minghui Chen$^{12}$ \enspace Meirui Jiang$^{3}$ \enspace Xin Zhang$^{4}$ \enspace Qi Dou$^{3}$ \enspace Zehua Wang$^{1}$ \enspace Xiaoxiao Li$^{12}$\thanks{Correspondence to \texttt{xiaoxiao.li@ece.ubc.ca}}  \\
  $^1$University of British Columbia \enspace $^2$Vector Institute \enspace $^3$Chinese University of Hong Kong \enspace $^4$Meta \\
}

\begin{document}
\maketitle
\begin{abstract}

Federated learning (FL) is a learning paradigm that enables collaborative training of models using decentralized data.
Recently, the utilization of pre-trained weight initialization in FL has been demonstrated to effectively improve model performance.
However, the evolving complexity of current pre-trained models, characterized by a substantial increase in parameters, markedly intensifies the challenges associated with communication rounds required for their adaptation to FL.
To address these communication cost issues and increase the performance of pre-trained model adaptation in FL, we propose an innovative model interpolation-based local training technique called ``Local Superior Soups.''
Our method enhances local training across different clients, encouraging the exploration of a connected low-loss basin within a few communication rounds through regularized model interpolation.
This approach acts as a catalyst for the seamless adaptation of pre-trained models in FL.
We demonstrated its effectiveness and efficiency across diverse widely-used FL datasets.
 Our code is available at \href{https://github.com/ubc-tea/Local-Superior-Soups}{https://github.com/ubc-tea/Local-Superior-Soups}.

\end{abstract}

\section{Introduction}
\label{sec:intro}

Federated learning (FL)~\cite{DBLP:conf/aistats/McMahanMRHA17} has emerged as a promising methodology for leveraging the power of private data without the need for centralized data governance.
However, data heterogeneity in FL poses significant challenges to the design of efficient training for global convergence.
With the emergence of the pre-training and fine-tuning paradigm in various applications~\cite{DBLP:conf/iccv/HeGD19, DBLP:conf/iclr/HuSWALWWC22}, recent studies~\cite{DBLP:journals/corr/abs-2210-08090, chen2022importance} have attempted to address the problem of FL under data heterogeneity with pre-trained initialization.
Although pre-trained federated learning can speed up convergence compared to random initialization, it still requires a significant number of communication rounds between the server and clients, often amounting to hundreds of rounds~\cite{DBLP:journals/corr/abs-2210-08090}.
Existing pre-trained models~\cite{DBLP:conf/icml/RadfordKHRGASAM21, DBLP:journals/corr/abs-2302-13971} often have an enormous parameter scale, and following the neural scaling law~\cite{DBLP:journals/corr/abs-2001-08361}, there is a continuous trend toward increasing model parameters. Deploying models with such a large parameter size in FL introduces significant communication overhead.
This greatly hampers the flexibility and scalability of model updates.
Reducing FL communication overhead can be approached by reducing the scale of model parameters involved in distributed training~\cite{DBLP:conf/acl/ZhangYDWYQX23} or reducing communication rounds~\cite{DBLP:conf/aistats/McMahanMRHA17}.
Comparing with reducing model parameters, reducing communication rounds typically leads to a more efficient reduction of network congestion~\cite{DBLP:conf/infocom/HegdeVM23}, decreased energy consumption on client devices~\cite{DBLP:journals/jsac/LuoLWHT21}, and a lower risk of privacy breaches~\cite{DBLP:conf/nips/ZhuLH19}. \textit{In this paper, we focus on reducing communication rounds in FL with pre-trained model as initialization. }

Typically, increasing the number of local training steps can effectively reduce communication rounds. However, there is an upper limit to the extent of local training step increments. This limitation arises due to the presence of data heterogeneity, where the consistency of optimization among different clients deteriorates with the increasing number of local steps~\cite{DBLP:conf/aistats/McMahanMRHA17}. This optimization inconsistency leads to a discrepancy between local and global models and decelerates the convergence rate of FL. The discrepancy is often called client drift~\cite{DBLP:conf/icml/KarimireddyKMRS20}.
Previously, some FL methods~\cite{DBLP:conf/icml/KarimireddyKMRS20, DBLP:conf/iclr/Sun0H0T23} attempted to introduce proximal terms to regularize local training, with the aim of reducing local overfitting and minimizing the problem of client drift.
While these methods can accelerate convergence, they restrict the progress of each local training steps towards the optimal solution, impeding the attainment of FL with more aggressive communication round reductions.
\begin{wrapfigure}{r}{0.5\textwidth}
    \centering
    \setlength{\abovecaptionskip}{0mm}
    \includegraphics[width=0.5\columnwidth]{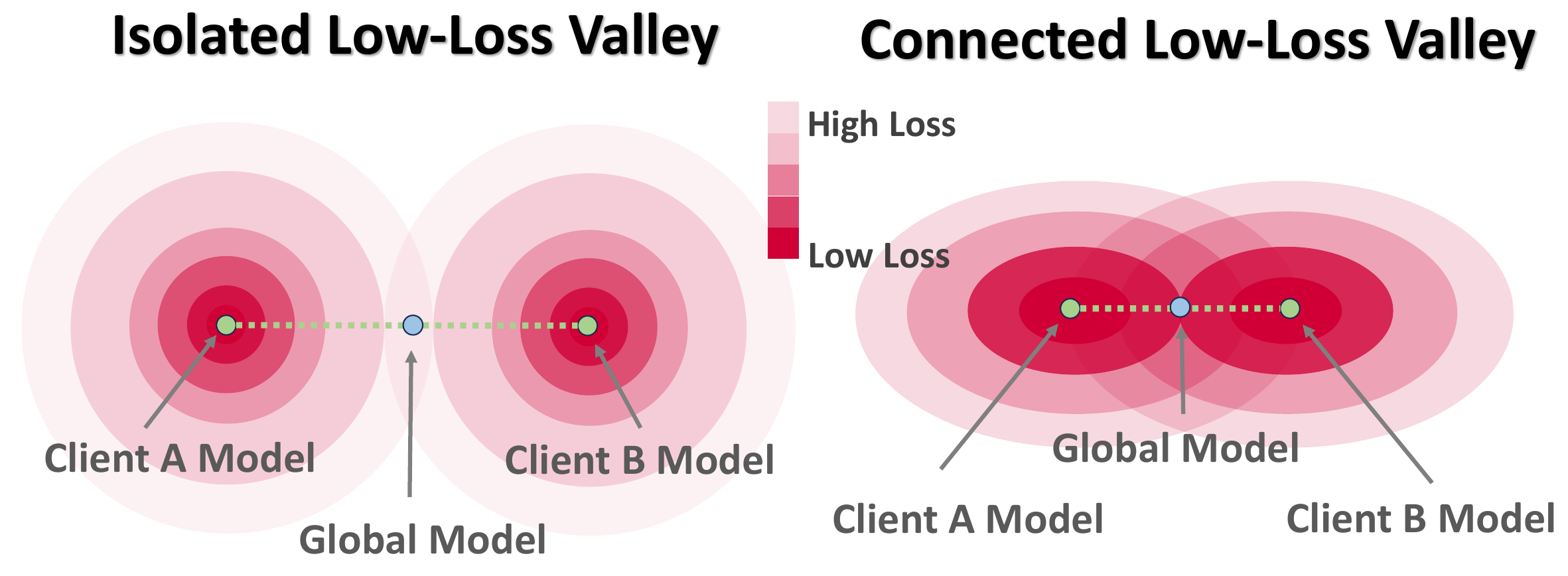}
    \caption{\small Illustration on isolated (left) and connected low-loss valley with larger regions in dark red (right).}
    \label{fig:loss_valley}
    \vspace{0mm}
\end{wrapfigure}

While these client drift mitigation methods can reduce local overfitting to some extent, they cannot ensure strong performance of the global aggregated models, particularly in scenarios with limited communication rounds.
This situation arises when individual local clients become trapped in isolated low-loss valleys. More specifically, as illustrated in Figure~\ref{fig:loss_valley}, two models from clients `A' and `B', even if their optimal model distance is small, still result in a poorly performing aggregated global model.
Moreover, the preceding FL methods aimed at minimizing communication rounds exclusively address scenarios involving random initialization, lacking a customized approach tailored to pre-trained models.
Recent proposed centralized fine-tuning methods (\eg, model soups~\cite{DBLP:conf/icml/WortsmanIGRLMNF22} and DiWA~\cite{DBLP:conf/nips/RameKRRGC22} -- a diverse weight-averaging approach) based on model interpolation (averages of a large number of model weights) are effective approaches to seek large connected low-loss region, which are promising for applying in FL to reduce communication rounds.
These methods can prevent individual clients from being trapped in isolated low-loss valleys
by positioning the global model centrally within a larger low-loss region by overlapping the low-loss regions among clients, as shown in Fig.~\ref{fig:loss_valley} (right).
However, their training efficiency is exceedingly low, requiring complete retraining of numerous models, leading to \emph{significant computational overhead} on clients and intolerable communication costs when applied in FL, due to two aspects:
First, they involve a time-consuming model selection phase within the model pool, which consists of all candidate models available for weight interpolation.
Secondly, model soups entail an extensive number of model training iterations, lacking prior guidance and relying on brute-force, random, and often redundant training. Many of the trained models end up unused.

To enjoy the connected low-loss valley benefits of model soup-based methods~\cite{DBLP:conf/icml/WortsmanIGRLMNF22, DBLP:conf/nips/RameKRRGC22} without burdening local training, we propose an efficient and local model interpolation-based method, called \textbf{Local Superior Soups (\ours{})}.  To address the first issue, we propose a \textbf{sequential random model interpolation} method during training. This eliminates the need for subsequent model selection steps and encourages low loss at sampled model interpolations during training
(Sec.~\ref{sec:random_interpolation}).
For the second issue, we introduce two quantifiable indicators of candidate model quality, inspired by data augmentation quality quantification~\cite{DBLP:journals/corr/abs-2002-08973, DBLP:conf/aaai/ChenWZH022}: \textbf{diversity} (Sec.~\ref{sec:diversity_term}) and \textbf{affinity} (Sec.~\ref{sec:affinity_term}).
Specifically, the \textit{diversity indicator} quantifies the diversity among models in the model pool with their pairwise model distance, where larger distances denote greater parameter diversity, used as a proxy for candidate non-redundancy.
\begin{wrapfigure}{r}{0.5\textwidth}
    \centering
    \setlength{\abovecaptionskip}{0mm}
    \includegraphics[width=0.5\columnwidth]{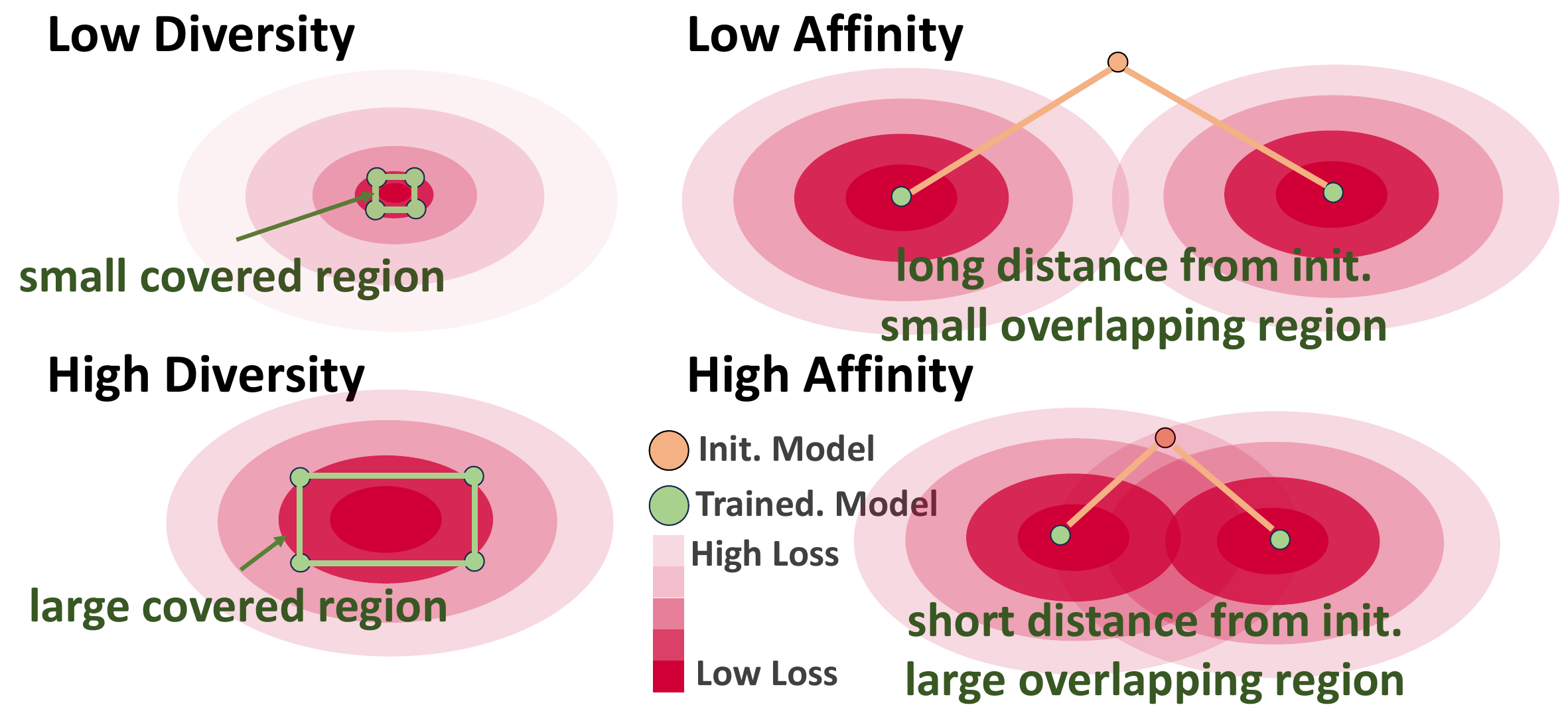}
    \caption{\small Illustration on diversity (left) and affinity (right) regularization.}
    \label{fig:lss_reg_term}
    \vspace{0mm}
\end{wrapfigure}

As illustrated in Figure~\ref{fig:lss_reg_term} (left), a low-loss region, supported by models with low diversity, can be effectively covered with only a few candidate models positioned near its periphery. Thus, we propose incorporating the diversity metric as a regularization term during training to maximize the expansion of low-loss regions, thereby increasing the utilization of trained models.
The \textit{affinity indicator} measures the affinity of each candidate model in the model pool to the initial model. %
Smaller distances indicate greater affinity. This affinity is also incorporated as a regularization term during training to prevent the expansion of low-loss regions from deviating too far from the shared initialization point, thus increasing the likelihood of overlapping connected regions (as depicted on the right side of Fig.~\ref{fig:lss_reg_term}). These two indicators facilitate the efficient inclusion of models into the model pool, preventing wasteful training of models that may ultimately go unused.
In experiments, we found that our proposed method greatly reduces communication rounds, and we achieved the performance of models fused after multiple rounds of communication in other FL methods with only a few rounds of communication.

In summary, our contributions are as follows.

(1) We reveal the importance of regularizing local client models in the connected low-loss valleys for reducing communication rounds when initializing FL with pre-trained models.
(2) We introduce an innovative and efficient model soups-based method for FL, called Local Superior Soups (\ours{}) that eliminates the need for time-consuming model selection and redundant model training in the existing soups-based approaches, while expanding connected low-loss valleys of client models for faster convergence.
(3) In experimental evaluations, \ours{} demonstrates a significant reduction in communication rounds, achieving superior performance with only a few rounds of communication, exceeding baseline FL methods significantly in four datasets and two types of distribution shifts.

\section{Related Work}

\subsection{Heterogeneous Federated Learning}

FL struggles with Non-IID data, leading to various proposed algorithms. FedProx \cite{DBLP:conf/mlsys/LiSZSTS20} uses proximal term to regularize local training, preventing client divergence. Scaffold \cite{DBLP:conf/icml/KarimireddyKMRS20} adds variance reduction to combat "clients-drift." MOON \cite{DBLP:conf/cvpr/LiHS21} employs mode-level contrastive learning to stabilize local training.
Personalized FL \cite{DBLP:journals/corr/abs-2103-00710} targets high local performance on Non-IID data. FedBN \cite{DBLP:conf/iclr/LiJZKD21} applies local batch normalization to mitigate feature shift before model averaging.
Recent one-shot and few-round FL methods use parallel server-side techniques like prediction ensembles~\cite{DBLP:journals/corr/abs-1902-11175}, data generation~\cite{DBLP:conf/nips/0081C0L0DS022, heinbaugh2023datafree}, or meta-learning~\cite{DBLP:conf/nips/ParkHKSM21} to improve aggregated model performance.

\subsection{Federated Fine-tuning and Model Interpolation}
\label{sec:related-ft}
Fine-tuning leverages pre-trained models to enhance task performance \cite{DBLP:journals/corr/abs-2211-00107}. FedFTG \cite{DBLP:conf/cvpr/ZhangS0TD22} proposes knowledge distillation for global model fine-tuning in FL. Personalized FL employs fine-tuning to adapt global models to local ones, \eg, FedBABU \cite{DBLP:journals/corr/abs-2106-06042}, FTFA, and RTFA \cite{DBLP:journals/corr/abs-2108-07313}. However, this focus on local performance neglects global generalization. Inspired by linear mode connectivity \cite{DBLP:conf/nips/NagarajanK19, DBLP:conf/icml/FrankleD0C20}, Model Soups \cite{DBLP:conf/icml/WortsmanIGRLMNF22} combines runs with varied hyper-parameters to improving fine-tuning performance in the centralized setting. DiWA \cite{DBLP:conf/nips/RameKRRGC22} and other Soups-based methods \cite{DBLP:conf/icml/WortsmanIGRLMNF22, DBLP:conf/nips/RameKRRGC22, DBLP:conf/miccai/ChenJDWL23} extends this concept, emphasizing the importance of model diversity. Some methods induce diversity through high learning rates \cite{DBLP:conf/nips/MaddoxIGVW19}, cosine similarity minimization \cite{DBLP:conf/icml/WortsmanHGFR21}, tempered posteriors \cite{DBLP:conf/uai/IzmailovMKGVW19}, or auxiliary dataset-trained model soups \cite{DBLP:journals/corr/abs-2212-10445}. We depict the difference of different model ensemble-based methods in our appendix~\ref{fig:soup_comparison}.

\section{Method}
\label{sec:method}

The structure of this Section is as follows: {firstly}, we provide the problem definition and corresponding notions to be used (Sec.~\ref{sec:notation}); {secondly}, we reveal the dilemma for existing federated learning methods on reducing communication rounds (Sec.~\ref{sec:tradeoff}); {finally}, we propose a regularized model interpolation-based method as a solution, provide corresponding analysis (Sec.~\ref{sec:solution}), and present the overall algorithm flow.

\subsection{Notions and Problem Definition}
\label{sec:notation}

\noindent\textbf{Notions.}
Let $\gX$ be the input space of data, $\gY$ be the label space.
Consider a FL setting with $\numClients$ clients, $\tau$ local steps and $R$ communication rounds.
Let $\mathcal{D}:=\left\{\mathcal{D}_{i}\right\}_{i=1}^{M}$ be a set of $M$ client datasets, each sampled from a distribution over $\mathcal{X}\times\mathcal{Y}$.
For each client, we have access to $n$ training data points in the form of $(\mathcal{X}_i, \mathcal{Y}_i) = \{(x_{j}^{i}, y_{j}^{i})\}_{j=1}^{n}$, where $y_{j}^{i}$ denotes the target label for input $x_{j}^{i}$.
Let $f \in \mathbb{R}^\modelSize$ represents the parameter for the global model, $\ell_i(f,\xi)$ denotes the per-sample loss at client $i$, and $\clientDist$ denotes a distribution on the entire set of clients.
We provide a notation table in Appendix~\ref{sec:appendix_notation} to clarify the meanings of the corresponding notations.

\noindent\textbf{Problem definition.}
We aim to address the challenge of optimizing the global performance on $\mathcal{D}$  of aggregated models fine-tuned from different clients with data heterogeneity, while minimizing communication rounds between the clients and the server in data Non-IID setting. In terms of global performance, we perform empirical risk minimization (ERM) on the sampled data $\mathcal{D}_i$ for $i \in [M]$,

\begin{align}
    \mathcal{L}(f) = \sum_{i=1}^\numClients p_i \mathcal{L}_i(f), \quad \text{where} \ \mathcal{L}_i(f)=\frac{1}{|\mathcal{D}_i|}\sum_{\xi \in \mathcal{D}_i}\ell_i(f,\xi) \ \text{and} \ \sum_{i=1}^\numClients p_i =1. \label{eqn:global_obj_erm}
\end{align}

\subsection{Effect of Regularization and Local Steps on FL Convergence under Data Heterogeneity}
\label{sec:tradeoff}

Following \cite{DBLP:journals/corr/abs-2107-06917}, we analyze convex local SGD under the assumptions in Appendix~\ref{app:theformal}.

\begin{theorem}[Conditional Convergence Rate for Perturbed Convex Local SGD]
    \label{thm:1}
    Under the full-participation, equal-weight convex local-SGD assumptions in Appendix~\ref{app:theformal}, with bounded gradient perturbations $c$ and $d>0$, choose
    $
    \eta = \min\{\frac{1}{4\beta}, \frac{M^{\frac{1}{2}}d}{\tau^{\frac{1}{2}} R^{\frac{1}{2}}\sigma}, \frac{d^{\frac{2}{3}}}{\tau^{\frac{2}{3}} R^{\frac{1}{3}} \beta^{\frac{1}{3}} \sigma^{\frac{2}{3}}}, \frac{d^{\frac{2}{3}}}{\tau R^{\frac{1}{3}} \beta^{\frac{1}{3}} (\zeta + c)^{\frac{2}{3}}}\} \\
    $
    we define $\epsilon = \Exs \left[  \frac{1}{\tau R} \sum_{r=0}^{R-1} \sum_{k=1}^{\tau} \mathcal{L}( \overline{f}^{(r,k)} ) - \mathcal{L}(f^{\star}) \right]$, and have
    \begin{equation}
        \begin{aligned}
        \epsilon \leq \frac{2 \beta d^2}{\tau R} + \frac{2 \sigma d}{\sqrt{M \tau R}} + \frac{5 \beta^\frac{1}{3} \sigma^\frac{2}{3} d^\frac{4}{3}}{\tau^\frac{1}{3} R^\frac{2}{3}} + \frac{19 \beta^\frac{1}{3} (\zeta + c)^\frac{2}{3} d^\frac{4}{3}}{R^\frac{2}{3}} + \mathcal{A}_{\mathrm{bias}} \\
        \label{eq:main_thm:3}
        \end{aligned}
    \end{equation}
\end{theorem}

The aggregate-bias remainder $\mathcal{A}_{\mathrm{bias}}$ is defined in Appendix~\ref{app:theformal} and need not vanish. The bound concerns averaged convex local-SGD iterates, rather than the final Adam soup.

Besides, $d := \|f^{(0,0)} - f^{\star}\|$ refers to the distance between initialization $f^{(0,0)}$ and the global optimum $f^{\star}$, $\sigma^2$ bounds stochastic-gradient variance by $\Exs [\|g_i(f^{(r,k)}) - \nabla \mathcal{L}_i(f^{(r,k)}) \|^2 | f^{(r,k)}] \leq \sigma^2$,
{and $\zeta$ bounds client--global gradient disagreement by $\max_{i} \sup_{f} \left\| \nabla \mathcal{L}_i(f) - \nabla \mathcal{L}(f) \right\| \leq \zeta.$}

\noindent\textbf{How to reduce communication rounds under data heterogeneity?}
At fixed $K=M\tau R$, the third and fourth terms in Eq.~\ref{eq:main_thm:3} grow as $\tau^{1/3}$ and $\tau^{2/3}$; saving rounds therefore requires controlling local steps.

\begin{proposition}\label{th:prop}
For the conditional local-SGD process of Theorem~\ref{thm:1}, fix $K=M\tau R$ and assume $\sigma,d,\beta>0$. If
\begin{align}\label{eqn:upb_tau}
1\le\tau\le\min\left\{\frac{\sigma\sqrt K}{d\beta M^2},\frac{\sigma}{\zeta+c}\sqrt{\frac{\sigma\sqrt K}{d\beta M^2}}\right\},
\end{align}
then the four explicit terms in Eq.~\ref{eq:main_thm:3} are $\mathcal{O}(\sigma d/\sqrt K)$. The full bound additionally retains $\mathcal{A}_{\mathrm{bias}}$. A term with zero denominator is omitted. See Appendix~\ref{app:proof} for the proof.
\end{proposition}

\noindent\textbf{Implications for pre-trained local training.}
Pre-training can improve client alignment~\cite{DBLP:journals/corr/abs-2210-08090} and, under suitable overparameterization assumptions, reduce the distance to an optimum~\cite{DBLP:conf/nips/JacotHG18, DBLP:conf/nips/ChizatOB19, DBLP:journals/corr/abs-2102-13522}.

\subsection{Our Solution: \ours{} Algorithm}
\label{sec:solution}

In this part, we first present the shortcomings of the previous model soups method applied in FL.
Secondly, we propose our three targeted improvements, \ie{} \textbf{\emph{ random model interpolation}} (Sec.~\ref{sec:random_interpolation}), \textbf{\emph{diversity term}} (Sec.~\ref{sec:diversity_term}), and \textbf{\emph{affinity regularization term}} (Sec.~\ref{sec:affinity_term}).
Finally, we present the complete algorithm process and detailed implementation in local client training.

\begin{wrapfigure}{R}{0.5\textwidth}
\vspace{-20pt} %
\begin{minipage}{0.5\textwidth}
\begin{algorithm}[H]
    \caption{\ours{} (Local Training) Pseudo-code} \label{alg:pseudo-code}
    \begin{algorithmic}[1]
        \REQUIRE $f_p$ pre-trained model (first round) or global aggregated model (later rounds); $\mathcal{L}$ loss function; $\mathcal{D}$ dataset; $dist$ distance function; $\tau$ iteration steps; $\eta$ learning rate; $\lambda_a$ affinity coefficient; $\lambda_d$ diversity coefficient; $N_c$ number of added candidate models.
        \STATE \textbf{\underline{\ours{} Local Training }:}
        \STATE $\gM \leftarrow \{f_p\}$
        \FOR{$i=1$ to $N_c$}
            \STATE ${f_{p_i}} \leftarrow Averaging(\gM)$
            \STATE $\gM \leftarrow \gM \cup \{{f_{p_i}}\}$ \algorithmiccomment{sequential training with newly added model}
            \FOR{$t=1$ to $\tau$}
                \STATE $f_s \leftarrow RandomInterpolation(\gM)$ \algorithmiccomment{connectivity preserving}
                \STATE $\mathcal{L}_{reg}({f_{p_i}}) = \mathcal{L}(f_s, \mathcal{D}) + \lambda_a \cdot dist({f_{p_i}}, f_p) - \lambda_d \cdot dist({f_{p_i}}, \gM)$
                \STATE ${f_{p_i}} \leftarrow {f_{p_i}} - \eta \nabla_{{f_{p_i}}} \mathcal{L}_{reg}({f_{p_i}})$
            \ENDFOR
        \ENDFOR
        \STATE \textbf{\underline{Inference}:}
        \STATE $f \leftarrow Averaging(\gM)$
    \end{algorithmic}
\end{algorithm}
\end{minipage}
\vspace{-10pt} %
\end{wrapfigure}

\noindent\textbf{Limitation of previous model soups methods.}
Previous model soups methods~~\cite{DBLP:conf/icml/WortsmanIGRLMNF22} can induce a trained model located in a connected low-loss valley, but their training efficiency is exceedingly low, due to two aspects:
\textit{Time-Consuming model selection phase:} Firstly, these methods involve a time-consuming model selection phase, which consists of all candidate models available for weight interpolation~\cite{DBLP:conf/miccai/ChenJDWL23, DBLP:conf/icml/WortsmanIGRLMNF22}. This phase aims to choose models that reside within the same low-loss valleys. During this selection process, significant computational resources are consumed to identify suitable models for interpolation, adding to the overall training time and complexity.
\textit{Extensive and redundant model training:} Secondly, model soups entail an extensive number of model training iterations, lacking prior guidance and relying on brute-force, random, and often redundant training~\cite{DBLP:journals/corr/abs-2309-15698, DBLP:conf/icml/WortsmanIGRLMNF22}. Many of the trained models end up unused, further exacerbating the computational inefficiency.

\subsubsection{Random interpolation conserving connected low-loss region.}
\label{sec:random_interpolation}
To address the \emph{time-consuming model selection} issue of the previous soups-based method, we propose a sequential random model interpolation method during training.
This innovative approach streamlines the training process by eliminating the need for subsequent model selection steps within the \textbf{model pool} (\ie, local models to be interpolated), which traditionally consumes a considerable amount of computational resources and time.
Let \( \mathcal{M} = \{ f_{p_1}, f_{p_2}, \ldots, f_{p_N} \} \) be a pool of \( N \) models, where \( f_{p_i} \) represents the weights of the \( i \)-th model.
We define the interpolated model \( f_{\text{interp}} \) as a weighted combination of the models in \( \mathcal{M} \). The interpolation coefficients \( \alpha = (\alpha_1, \alpha_2, \ldots, \alpha_N) \) are sampled using a uniform distribution and normalization strategy.
The interpolated model \( f_{\text{interp}} \) is then computed as:
$f_{\text{interp}} = \sum_{i=1}^{N} \alpha_i f_{p_i}$.
Here, \( \alpha_i \) represents the weight assigned to the \( i \)-th model in the pool. Normalization ensures that the non-negative coefficients \( \alpha_i \) sum to 1, providing a simple and effective way to combine the model weights from the pool \( \mathcal{M} \).
Forward and backward propagation are performed using the interpolated model, updating the weights of the currently active model (\ie, the newly added model ) (corresponding to Algorithm~\ref{alg:pseudo-code} Line $7$), while previously added model weights remain fixed.

\subsubsection{Diversity term.}
\label{sec:diversity_term}
The diversity term is proposed to address the \emph{redundant model training} issue of the previous soups-based methods by encouraging low-loss region expansion.
In particular, the diversity indicator assesses the variability among models within the model pool by summing the distances between pairs of models. Greater distances between models indicate a higher degree of parameter diversity, used as a proxy for candidate non-redundancy.
This diversity metric is integrated into the FL local training process as a regularization term to facilitate the extensive enlargement of low-loss regions, consequently maximizing the effectiveness of trained models.
The diversity term (in Algorithm~\ref{alg:pseudo-code} Line 8)  measures the distance between the current training model and other models that will be averaged, and we hope that this distance to be large.
The diversity loss can be defined as
\begin{equation}
    \ell_\text{diversity} = \textit{dist}(f,\gM) = \frac{1}{N} \sum_{n=1}^{N} \textit{dist}(f, f_{n}).
    \label{eq:d_dist}
\end{equation}
Here, $f_{n}$ belongs to local interpolated model pool $\gM$ and $N$ is the number of local candidate models. The candidate models (\ie, model soups ingredients) are models to be interpolated in local training, and the model pool is the set of local candidate models (see Algorithm~\ref{alg:pseudo-code} Line 5). We use the $\ell_2$ norm to measure the distance between model weights.

\subsubsection{Affinity term.}
\label{sec:affinity_term}
The affinity term is proposed to control the expansion of low-loss regions and prevent local candidate model training divergence.
The affinity indicator assesses the level of alignment between each candidate model within the model pool and the initial global model by calculating the cumulative distances between each candidate model and the initialization model. Smaller distances between models signify a stronger affinity. To ensure the controlled expansion of low-loss regions and increase the probability of overlapping connected regions, this affinity metric is integrated into the training process as a regularization term.
The affinity term (in Algorithm~\ref{alg:pseudo-code} Line 8) measures the distance between the candidate model and the initial model weights, with the aim of minimizing this dissimilarity to ensure that the distance remains relatively small.
The affinity loss can be defined as
\begin{equation}
    \ell_\text{affinity} = \textit{dist}(f, f_p).
    \label{eq:v_dist}
\end{equation}
Here, $f_{p}$ is the shared round initialization: the pre-trained model in the first round, and the global aggregated model in later rounds.
Moreover, it encourages each local model to lie in a close zone in the parameter space, which is beneficial for subsequent server aggregation, especially under data heterogeneity. We use $\ell_2$ distance for the $dist(\cdot,\cdot)$ metric for both Eq.~\ref{eq:d_dist} and  Eq.~\ref{eq:v_dist}.

\subsubsection{Overall pipeline.}
\label{sec:overall_pipeline}
We outline \ours{} as follows:
We begin with the initialization of the client's local model with the current global model. Then we will refine the local model using affinity and diversity loss.
This step is performed for a few local update steps. Finally, after updating local model, we aggregate them in the server following the common averaging operation in FedAvg~\cite{DBLP:conf/aistats/McMahanMRHA17}. The flow of \ours{} for local updating (Step 2 described in Sec~\ref{sec:notation}) can be found in Algorithm \ref{alg:pseudo-code}.

In conclusion, our method aims to minimize the distance between the local fine-tuned model and the current initialized global model while maximizing the distance between the model soups ingredients (\ie, the models to be averaged).
Our fine-tuned models find large low-loss regions on their respective local datasets while encouraging parameters close to the current initialization. It is intuitive that the parameters of our fine-tuned models can be more easily aligned with those of models fine-tuned on similar datasets, thereby improving communication efficiency.

\begin{table}[t]
    \centering
    \caption{\small Label shift test accuracy after $R = 1$ and $R = 3$ communication rounds. We primarily compared two categories of methods: conventional FL methods and state-of-the-art local weight averaging-based fine-tuning methods that enhance domain generalization.
    }
    \label{table:ls_main}
    \renewcommand\arraystretch{1.0}
    \resizebox{0.9\columnwidth}{!}{
    \begin{tabular}{l*8c}
        \toprule
        & \multicolumn{2}{c}{\textbf{FMNIST}} & \multicolumn{2}{c}{\textbf{CIFAR-10}} \\
        \cmidrule(lr){2-3} \cmidrule(lr){4-5}
        Method & Accuracy ($R = 1$) $\uparrow$  &  Accuracy ($R = 3$) $\uparrow$ &  Accuracy ($R = 1$) $\uparrow$ &  Accuracy ($R = 3$) $\uparrow$ \\
        \midrule
        FedAvg \cite{DBLP:conf/aistats/McMahanMRHA17} & $35.54 (1.71) $ & $90.04 (0.32) $ & $58.34 (0.86) $ & $66.74 (0.76) $ \\
        FedProx \cite{DBLP:conf/mlsys/LiSZSTS20} & $33.48 (1.52) $ & $89.28 (0.36) $ & $56.74 (0.92) $ & $63.21 (0.83) $ \\
        MOON \cite{DBLP:conf/cvpr/LiHS21} & $36.01 (1.66) $ & $91.28 (0.30) $ & $58.96 (1.24) $ & $67.04 (1.12) $ \\
        FedBN \cite{DBLP:conf/iclr/LiJZKD21} & $34.20 (1.73) $ & $89.87 (0.47) $ & $57.04 (0.75) $ & $64.51 (0.67) $ \\
        FedFomo \cite{DBLP:conf/iclr/ZhangSFYA21}  & $33.94 (1.65) $ & $88.41 (0.69) $ & $55.01 (0.89) $ & $62.69 (0.75) $ \\
        FedRep \cite{DBLP:conf/icml/CollinsHMS21}  & $36.20 (1.52) $ & $91.07 (0.23) $ & $57.73 (0.82) $ & $66.23 (0.73) $ \\
        FedBABU \cite{DBLP:journals/corr/abs-2106-06042}  & $36.18 (1.43) $ & $91.31 (0.26) $ & $60.14 (1.06) $ & $67.16 (0.87) $ \\
        \hline
        SWA \cite{DBLP:conf/uai/IzmailovPGVW18} & $55.82 (1.02) $ & $91.03 (0.19) $ & $59.07 (1.28) $ & $67.45 (1.15) $ \\
        SWAD \cite{DBLP:conf/nips/ChaCLCPLP21} & $58.66 (0.87) $ & $91.22 (0.16) $ & $60.54 (1.15) $ & $67.65 (0.97) $ \\
        Soups \cite{DBLP:conf/icml/WortsmanIGRLMNF22} & $60.11 (0.64) $ & $91.56 (0.24) $ & $61.00 (1.04) $ & $67.63 (0.94) $ \\
        DiWA \cite{DBLP:conf/nips/RameKRRGC22} & $63.21 (0.54) $ & $91.88 (0.13) $ & $61.32 (1.26) $ & $68.05 (1.10) $ \\
        \hline
        \textbf{\ours{}} (4x Mem.) & $\textbf{72.66} (0.73) $ & $\textbf{92.45} (0.21) $ & $\textbf{65.96} (1.50) $ & $\textbf{75.16} (1.07) $ \\
        \bottomrule
    \end{tabular}}
    \vspace{-5mm}
\end{table}

\begin{table}[t]
    \centering
    \caption{\small Feature shift test accuracy after $R = 1$ and $R = 3$ communication rounds. \ours{} consistently outperforms other methods on both datasets across under feature shift settings.}
    \label{table:fs_main}
    \renewcommand\arraystretch{1.0}
    \resizebox{0.9\columnwidth}{!}{
    \begin{tabular}{l*8c}
        \toprule
        & \multicolumn{2}{c}{\textbf{Digit-5}} & \multicolumn{2}{c}{\textbf{DomainNet}} \\
        \cmidrule(lr){2-3} \cmidrule(lr){4-5}
        Method & Accuracy ($R = 1$) $\uparrow$  &  Accuracy ($R = 3$) $\uparrow$ &  Accuracy ($R = 1$) $\uparrow$ &  Accuracy ($R = 3$) $\uparrow$ \\
        \midrule
        FedAvg \cite{DBLP:conf/aistats/McMahanMRHA17} & $46.36 (2.08) $ & $80.48 (0.81) $ & $18.76 (3.52) $ & $29.43 (2.01) $ \\
        FedProx \cite{DBLP:conf/mlsys/LiSZSTS20} & $ 44.01 (1.92) $ & $77.83 (0.68) $ & $17.27 (3.22) $ & $27.18 (2.29) $ \\
        MOON \cite{DBLP:conf/cvpr/LiHS21} & $50.11 (1.72) $ & $83.02 (0.64) $ & $19.61 (3.54) $ & $31.27 (2.34) $ \\
        FedBN \cite{DBLP:conf/iclr/LiJZKD21} & $46.02 (1.93) $ & $81.42 (0.71) $ & $18.16 (3.09) $ & $28.65 (1.89) $ \\
        FedFomo \cite{DBLP:conf/iclr/ZhangSFYA21}  & $41.87 (2.13) $ & $76.21 (0.98) $ & $15.10 (3.82) $ & $25.69 (2.38) $ \\
        FedRep \cite{DBLP:conf/icml/CollinsHMS21}  & $47.43 (1.73) $ & $82.02 (0.63) $ & $18.89 (2.60) $ & $30.42 (1.84) $ \\
        FedBABU \cite{DBLP:journals/corr/abs-2106-06042}  & $48.02 (1.81) $ & $83.20 (0.79) $ & $19.44 (2.43) $ & $32.06 (1.88) $ \\
        \hline
        SWA \cite{DBLP:conf/uai/IzmailovPGVW18} & $54.13 (0.72) $ & $85.33 (0.62) $ & $22.07 (2.55) $ & $35.90 (1.61) $ \\
        SWAD \cite{DBLP:conf/nips/ChaCLCPLP21} & $57.02 (0.71) $ & $86.84 (0.64) $ & $21.98 (2.61) $ & $36.73 (1.57) $ \\
        Soups \cite{DBLP:conf/icml/WortsmanIGRLMNF22} & $59.71 (0.82) $ & $87.07 (0.58) $ & $22.75 (2.85) $ & $38.02 (1.40) $ \\
        DiWA \cite{DBLP:conf/nips/RameKRRGC22} & $61.54 (0.83) $ & $88.83 (0.69) $ & $24.88 (2.54) $ & $38.32 (1.50) $ \\
        \hline
        \textbf{\ours{}} (4x Mem.) & $\textbf{72.86} (1.64) $ & $\textbf{92.97} (0.65) $ & $\textbf{27.86} (2.85) $ & $\textbf{41.35} (1.46) $ \\
        \bottomrule
    \end{tabular}}
    \vspace{-5mm}
\end{table}

\section{Experiment}
\label{sec:experiment}

\subsection{Experimental Setup}

\noindent\textbf{Dataset.}
Our experimental section considers two scenarios of Non-IID settings, namely label shift and feature shift. The label shift scenario investigates datasets such as FMNIST~\cite{DBLP:journals/corr/abs-1708-07747} and CIFAR10~\cite{krizhevsky2009learning}, while feature shift involves Digit5 and DomainNet. Further information on the specific datasets can be found in the appendix. In the label shift scenario, we partitioned the dataset into five clients and the data for each client are sampled following Dirichlet distributions with coefficient $\alpha = 1.0$, yielding imbalanced label distributions. In the feature shift scenario, we utilized five clients for Digit5~\cite{DBLP:conf/icml/GaninL15, DBLP:conf/iclr/LiJZKD21} and five clients for DomainNet~\cite{DBLP:conf/iccv/PengBXHSW19}. Additional results on an extended number of clients are presented in the appendix.

\noindent\textbf{Model.}
In terms of models, we used the ImageNet pre-trained ResNet50 \cite{DBLP:conf/cvpr/HeZRS16} as the base model for the DomainNet dataset, while for other datasets, we used the pre-trained ResNet-18 trained on ImageNet \cite{DBLP:conf/cvpr/DengDSLL009}. We also present the experimental results based on the vision transformer (ViT) model \cite{DBLP:conf/iclr/DosovitskiyB0WZ21} with parameter-efficient fine-tuning.

\noindent\textbf{Baselines.}
We compare \ours{} against the vanilla FL method - FedAvg~\cite{DBLP:conf/aistats/McMahanMRHA17} and several advanced FL algorithms designed for Non-IID settings, including FedProx~\cite{DBLP:conf/mlsys/LiSZSTS20}, MOON~\cite{DBLP:conf/cvpr/LiHS21}, FedBN~\cite{DBLP:conf/iclr/LiJZKD21}, FedFomo~\cite{DBLP:conf/iclr/ZhangSFYA21}, FedRep~\cite{DBLP:conf/icml/CollinsHMS21} and FedBABU~\cite{DBLP:journals/corr/abs-2106-06042}. Additionally, we make comparisons with top-performing weight/model-averaging-based domain generalization methods including SWA~\cite{DBLP:conf/uai/IzmailovPGVW18}, SWAD~\cite{DBLP:conf/nips/ChaCLCPLP21}, Soups~\cite{DBLP:conf/icml/WortsmanIGRLMNF22} and DiWA~\cite{DBLP:conf/nips/RameKRRGC22} by adapting them to FL.
In particular, the specific approach is to modify the local client training in the FedAvg framework to a corresponding fine-tuning approach. For more details, please refer to the appendix.

\noindent\textbf{Evaluation and implementation details.}
Unless otherwise specified, the model performance in the experiments below refers to the global model performance after aggregation on the server side.
Our training optimizer uses the Adam optimizer with a learning rate of $5\mathrm{e}{-4}$ and a training batch size of $64$.
For commonly used FL methods, due to the significant increase in local update steps that leads to worse convergence, we set their local update steps to $8$. For SWA, SWAD, and our method, we take more local update steps, with each model being averaged trained $8$ steps, and the default number of models to be averaged is $4$.
For the Model Soups method and DiWA, we train $32$ models with $8$ steps.
Additional details of experiment implementations are included in the Appendix.

\begin{figure}[ht]
    \centering
    \setlength{\abovecaptionskip}{-0.8mm}
    \subfloat[] {
        \label{fig:sub:cifar_head_comm}
        \includegraphics[width=0.22\columnwidth]{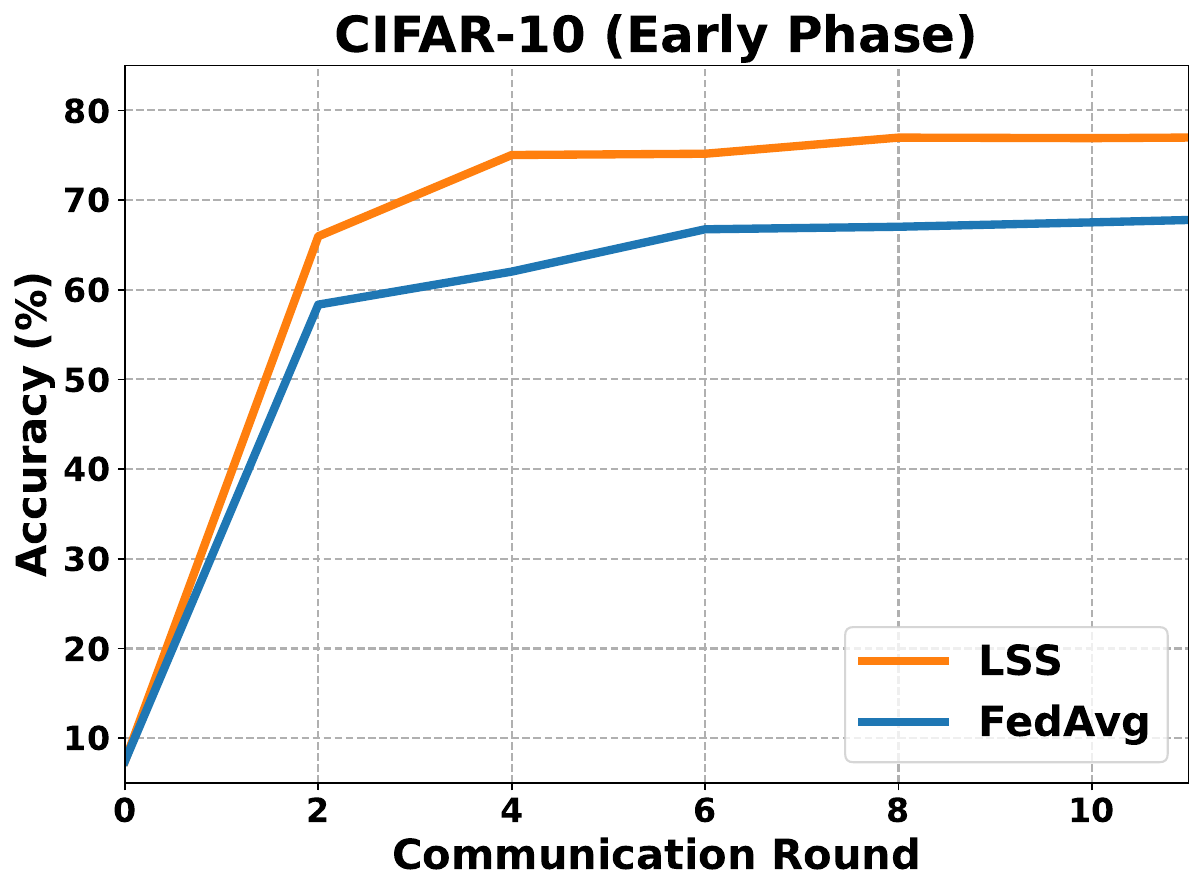}
    }
    \subfloat[] {
        \label{fig:sub:cifar_tail_comm}
        \includegraphics[width=0.22\columnwidth]{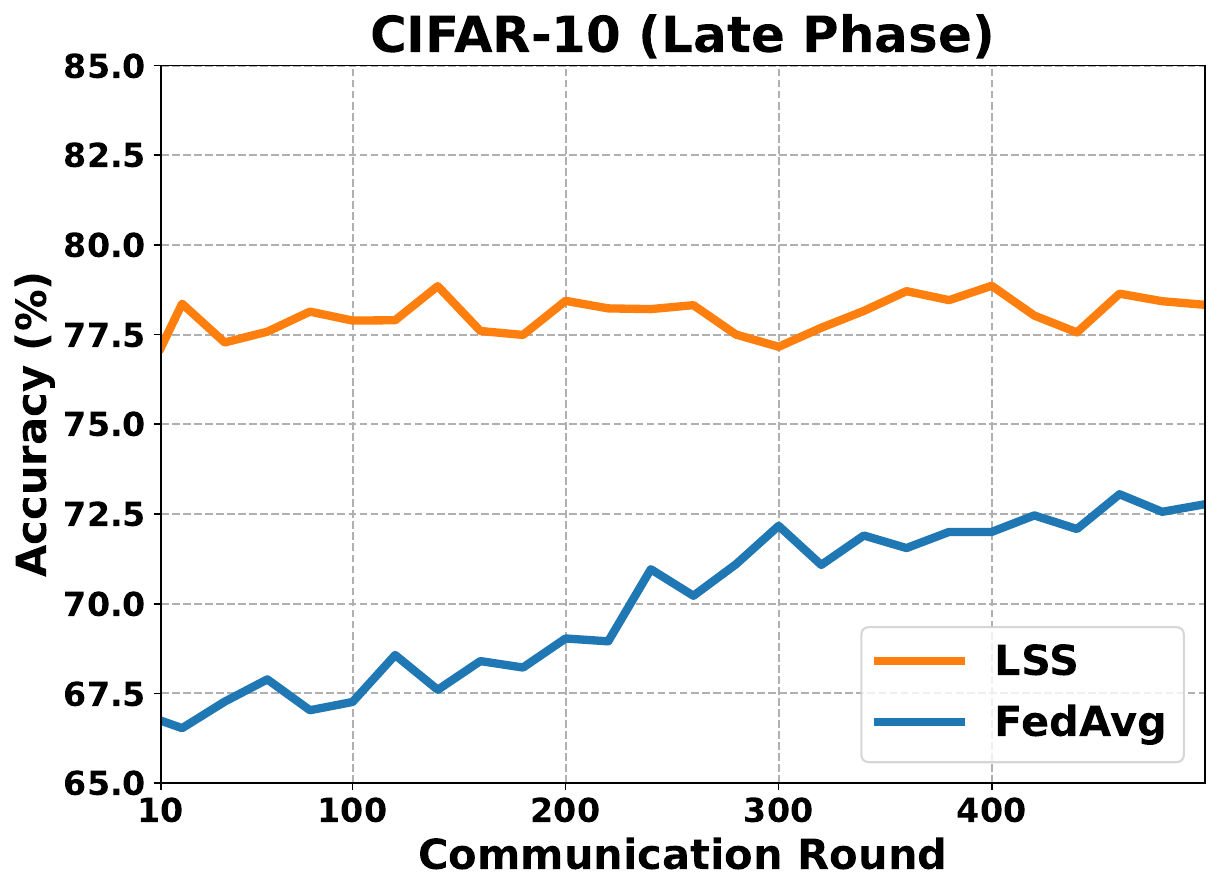}
    }
    \subfloat[] {
        \label{fig:sub:digit_head_comm}
        \includegraphics[width=0.22\columnwidth]{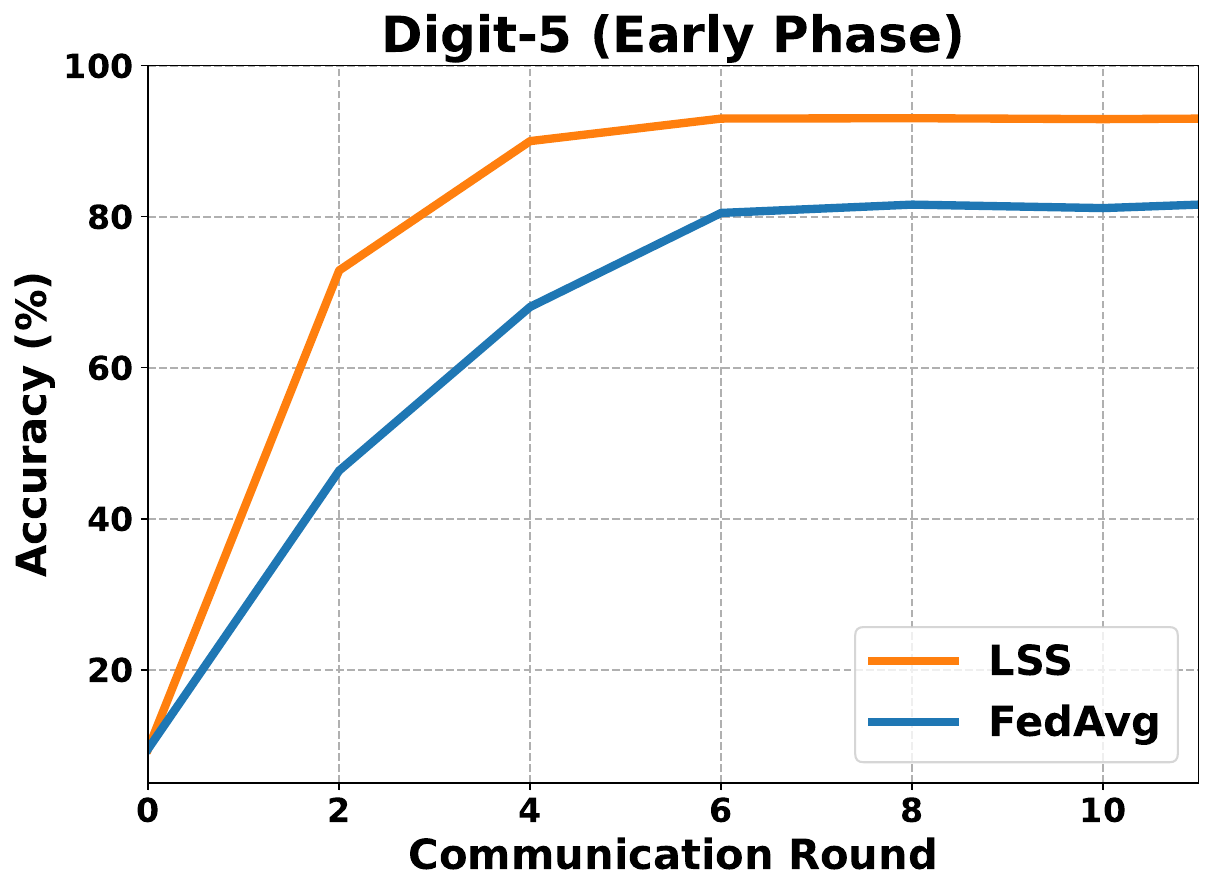}
    }
    \subfloat[] {
        \label{fig:sub:digit_tail_comm}
        \includegraphics[width=0.22\columnwidth]{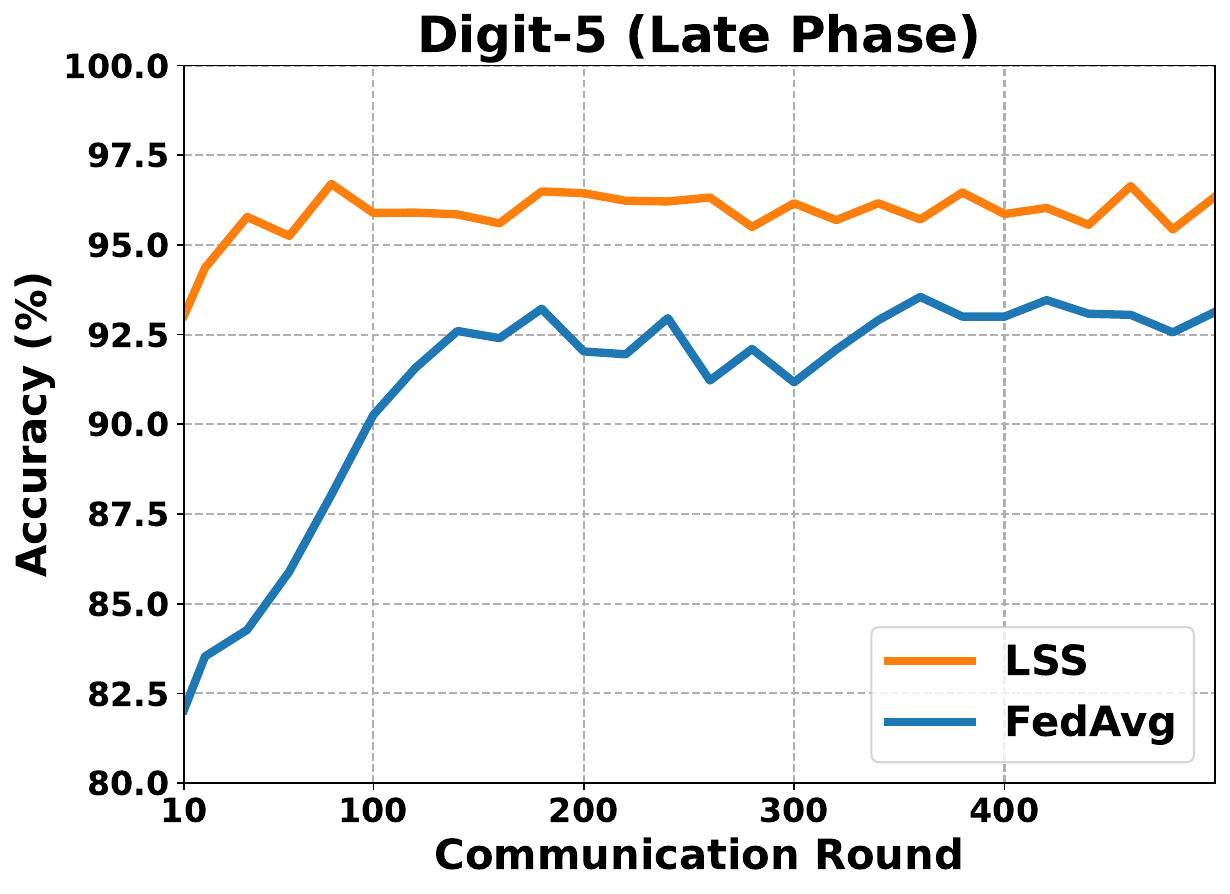}
    }
    \caption{\small Convergence comparison of our proposed \ours{} with FedAvg. \ours{} achieves high accuracy much earlier (around 6 to 8 rounds) than FedAvg, which takes hundreds of communication rounds.}
    \label{fig:convergence_speed}
    \vspace{-5mm}
\end{figure}

\subsection{Performance Comparison}
\noindent\textbf{Results on label shift.} To demonstrate the effectiveness of \ours{} on label shift scenario, we conduct comparison experiments on FMNIST and CIFAR-10 datasets. We consider fine-tuning with an extremely limited number of communication rounds (\ie, $R=1$ and $R=3$). Table~\ref{table:ls_main} reports the test accuracy with the format of mean (std) for all compared algorithms. All experiments are repeated $3$ runs with different random seeds. In Table~\ref{table:ls_main}, \ours{} achieves the best accuracy on all settings of both datasets, which validates that \ours{} is efficient and effective in fine-tuning FL for label shift Non-IID. Notably, on FMNIST with just one round of communication, \ours{} can double the accuracy of the best Non-IID FL baseline method. Surprisingly, the simple extension of model-averaging-based domain generalization methods onto FedAvg~\cite{DBLP:conf/aistats/McMahanMRHA17} (the 2nd big row in Table~\ref{table:ls_main}) perform very well, especially when the number communication round is small. The superior performance is consistent with the usefulness of local weight averaging in few-round fine-tuning. We further provide results on different levels of label shift in the supplementary material.

\noindent\textbf{Results on feature shift.} Table~\ref{table:fs_main} evaluates on feature shift scenario using Digits-5 and DomainNet datasets. Similar to the previous experiment setting for Table~\ref{table:ls_main}, we repeat all the algorithms with $3$ random seeds. Consistent with the observation in Table~\ref{table:fs_main}, \ours{} is the top-performing method under all the settings for both datasets. We also observe better performance achieved by adapting model-averaging-based domain generalization methods (the 2nd big row in Table~\ref{table:fs_main}) in FL than the existing Non-IID FL methods (the 1st big row in Table~\ref{table:fs_main}), which further verifies the effectiveness of model averaging to obtain better global model while improving communication efficiency.

\noindent\textbf{Convergence plots.} We also evaluate the strength of \textit{faster convergence} using the proposed \ours{} compared with FedAvg~\cite{DBLP:conf/aistats/McMahanMRHA17} on CIFAR-10 (label shift) and Digtis-5 (feature shift). Fig.~\ref{fig:convergence_speed} depicts the testing accuracies at early and late phases regarding the number of communication rounds to reach convergence. First, by looking at the final testing accuracies on Fig.~\ref{fig:convergence_speed} (b) and (d), \ours{} achieves better performance. Second, Fig.~\ref{fig:convergence_speed} (a) and (c) show that \ours{} almost meets the targeted performance at the very early stage (\ie around 6 to 8 rounds), whereas FedAvg requests over hundreds of communication rounds.

\begin{figure}[!htbp]
    \centering
    \setlength{\abovecaptionskip}{2mm}
    \subfloat[] {
        \label{fig:sub:peft_soup_r1}
        \includegraphics[width=0.35\columnwidth]{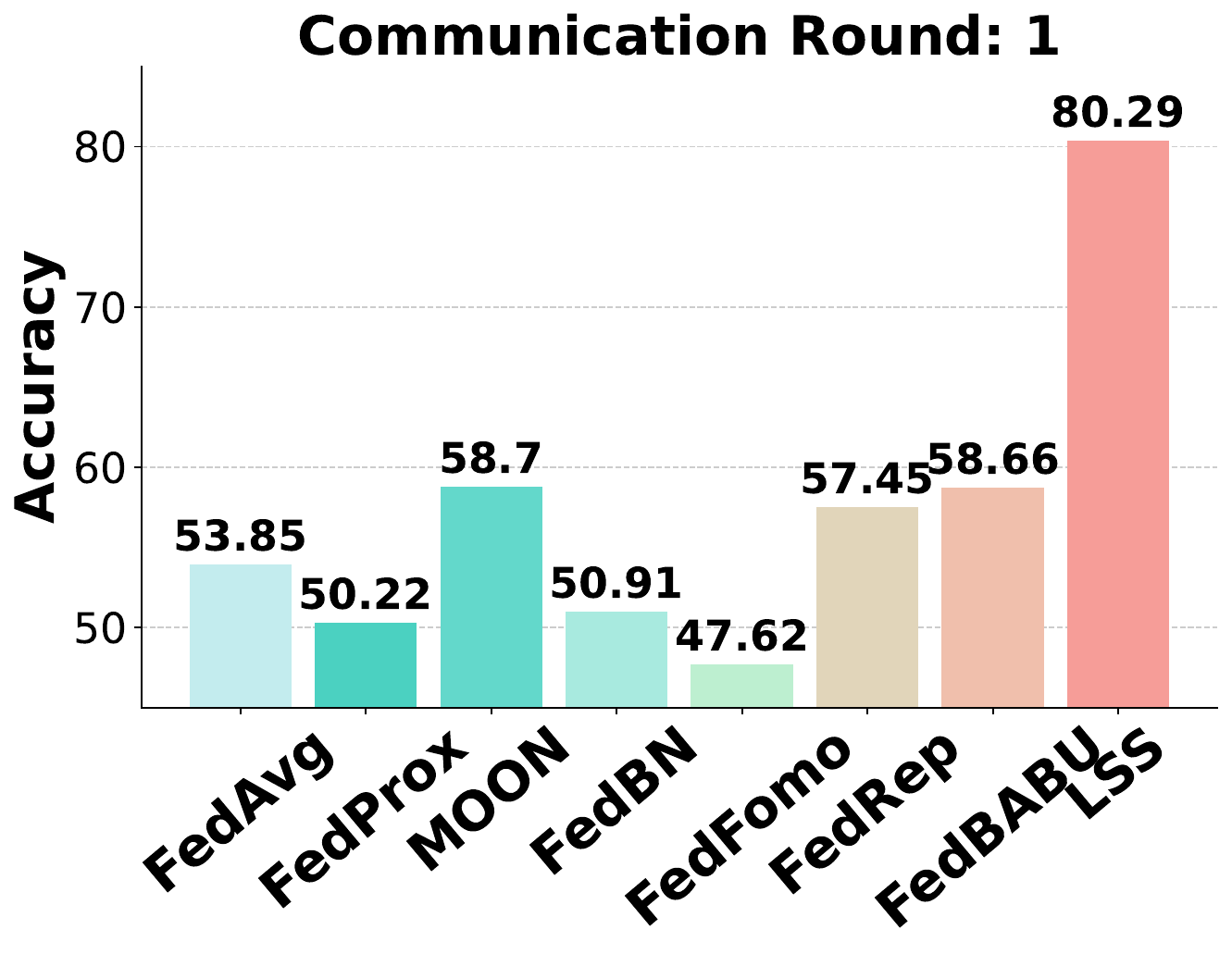}
    }
    \subfloat[] {
        \label{fig:sub:peft_soup_r3}
        \includegraphics[width=0.35\columnwidth]{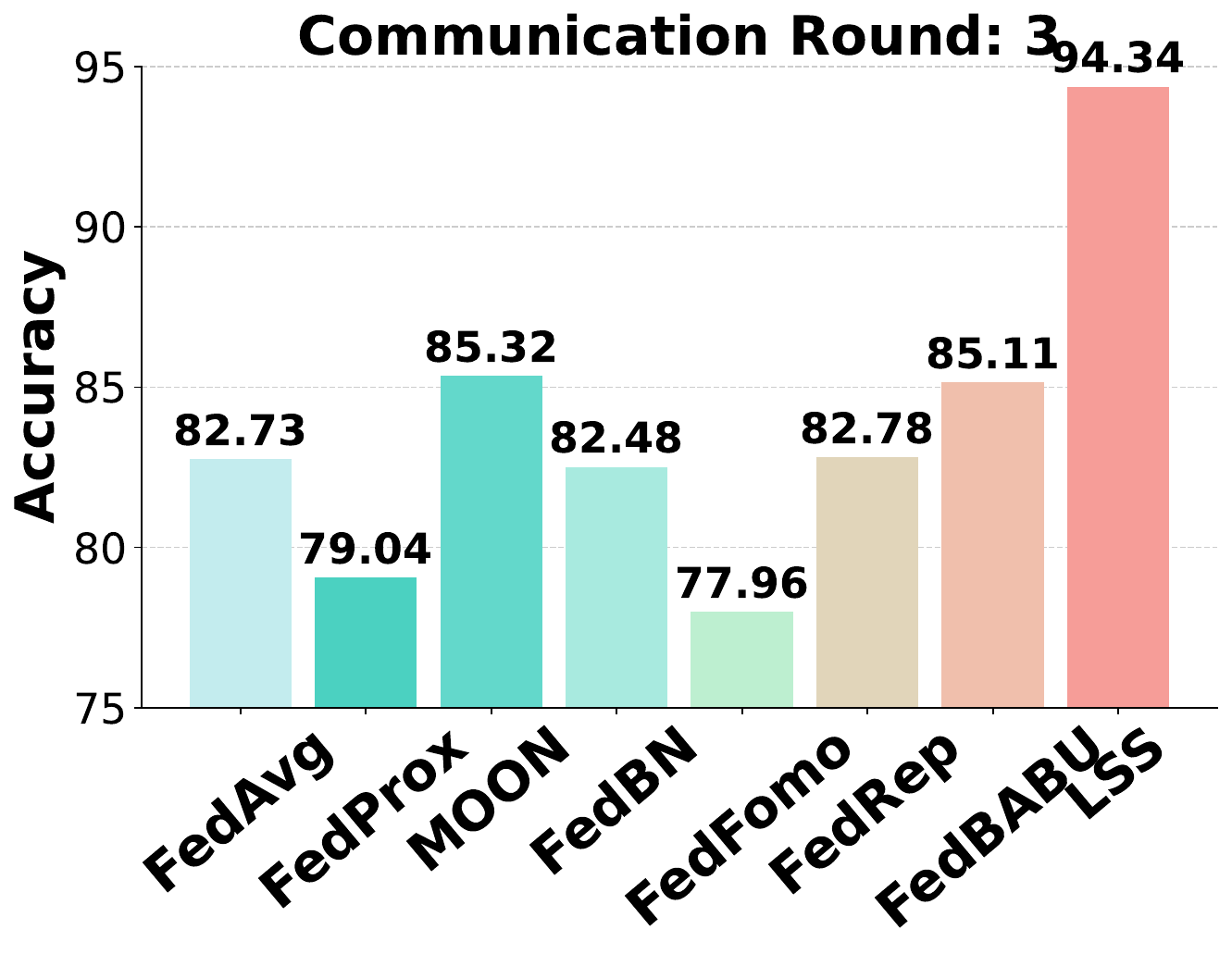}
    }
    \caption{Evaluation on ViT fine-tuned with LoRA (Digit5 dataset).
    }
    \label{fig:vit_lora_digit5}
    \end{figure}

\noindent\textbf{Parameter-Efficient Tuning with ViT.} We also deployed the Vision Transformer (ViT)~\cite{DBLP:conf/iclr/DosovitskiyB0WZ21} in FL learning. On Digits-5 dataset, we evaluate the ViT model with a resolution of $224$ and a patch size of $16$, which was pretrained on the ImageNet-21k dataset. Due to the large number of parameters in ViT, we used a parameter-efficient fine-tuning method called LoRA~\cite{DBLP:conf/iclr/HuSWALWWC22} to train it for all the methods. For more details about our ViT architecture and LoRA training, please refer to the appendix. It can be observed in Fig.~\ref{fig:vit_lora_digit5} that our method is applicable to pre-trained ViT models, demonstrating that our approach can be combined with parameter-efficient fine-tuning methods to further enhance the communication efficiency of FL.

\subsection{Ablation Studies}

We conducted ablation experiments on the main components (\textit{i.e.}, affinity, diversity term and averaged model quantity) of our proposed method and evaluated their performance on the CIFAR dataset, with the performance metric being the global model performance at communication round $R = 1$.

\begin{figure}[!htbp]
    \centering
    \setlength{\abovecaptionskip}{2mm}
    \subfloat[] {
        \label{fig:sub:aff_scatter}
        \includegraphics[width=0.35\columnwidth]{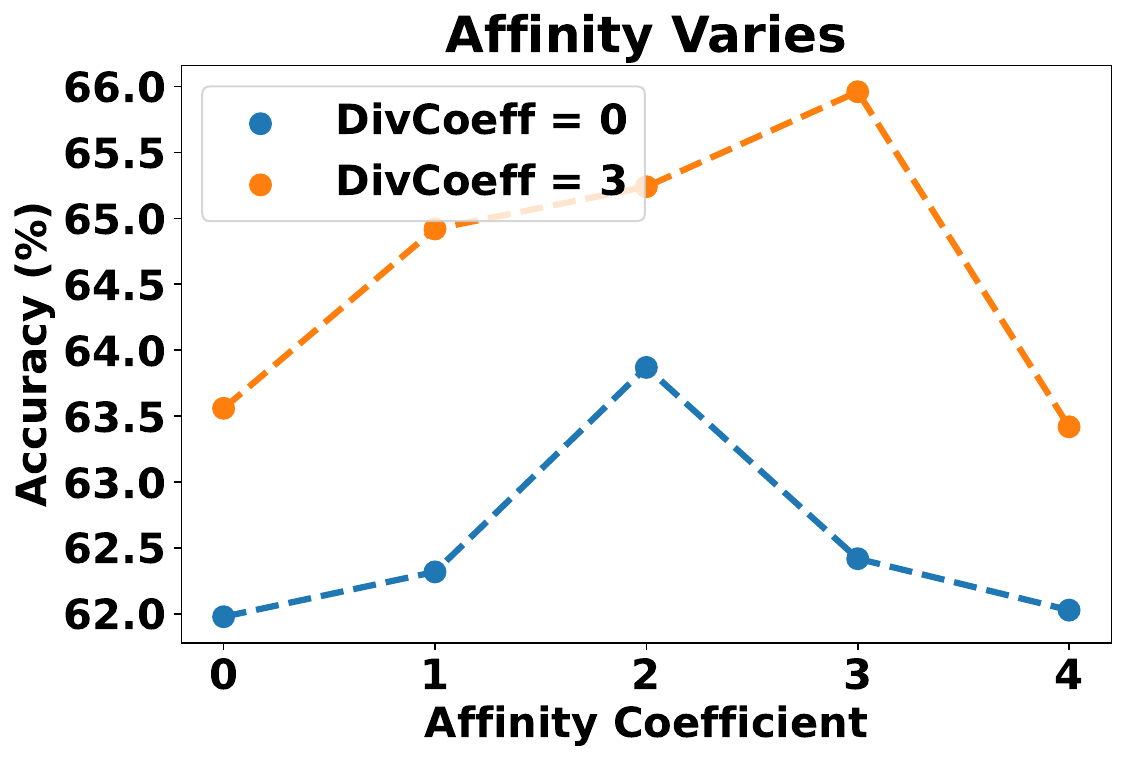}
    }
    \subfloat[] {
        \label{fig:sub:div_scatter}
        \includegraphics[width=0.35\columnwidth]{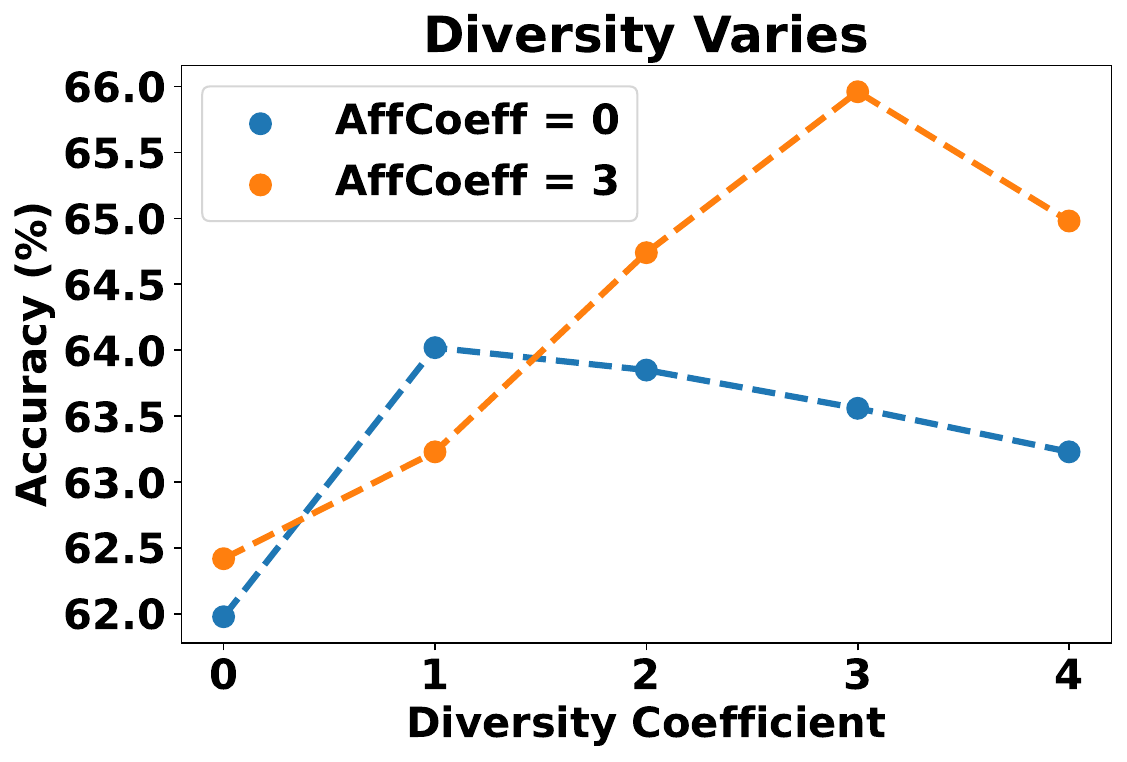}
    }
    \caption{Ablation on the affinity \& diversity.
    }
    \label{fig:aff_div}
    \end{figure}

\noindent\textbf{Investigation on regularization losses.} In order to examine the importance of affinity loss and diversity loss, as well as the influence of their corresponding coefficients, we adjust one coefficient within a range of 0 to 4 while maintaining the other at a constant value.
By comparing the performance with and without loss term, we observe that adding affinity and diversity terms can enhance the model's performance.
Additionally, we observe that the two terms complement each other, and selecting appropriate coefficients can achieve significant performance improvement (\eg, adjusting the affinity coefficient to $3$ as shown in Fig.~\ref{fig:aff_div} (a) and diversity coefficient to 3 as shown in Fig.~\ref{fig:aff_div} (b)).

\begin{figure}[!htbp]
    \centering
    \setlength{\abovecaptionskip}{2mm}
    \subfloat[] {
        \label{fig:sub:num_model_global}
        \includegraphics[width=0.32\columnwidth]{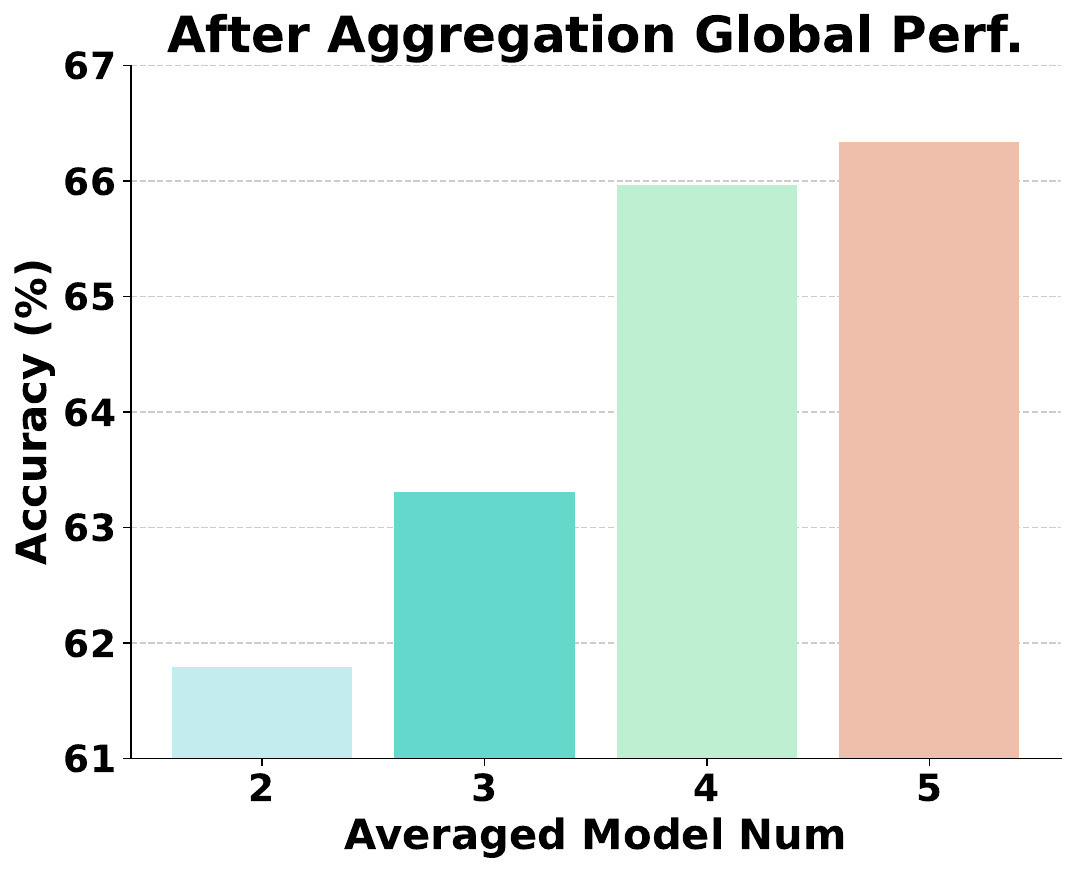}
    }
    \subfloat[] {
        \label{fig:sub:num_model_local}
        \includegraphics[width=0.32\columnwidth]{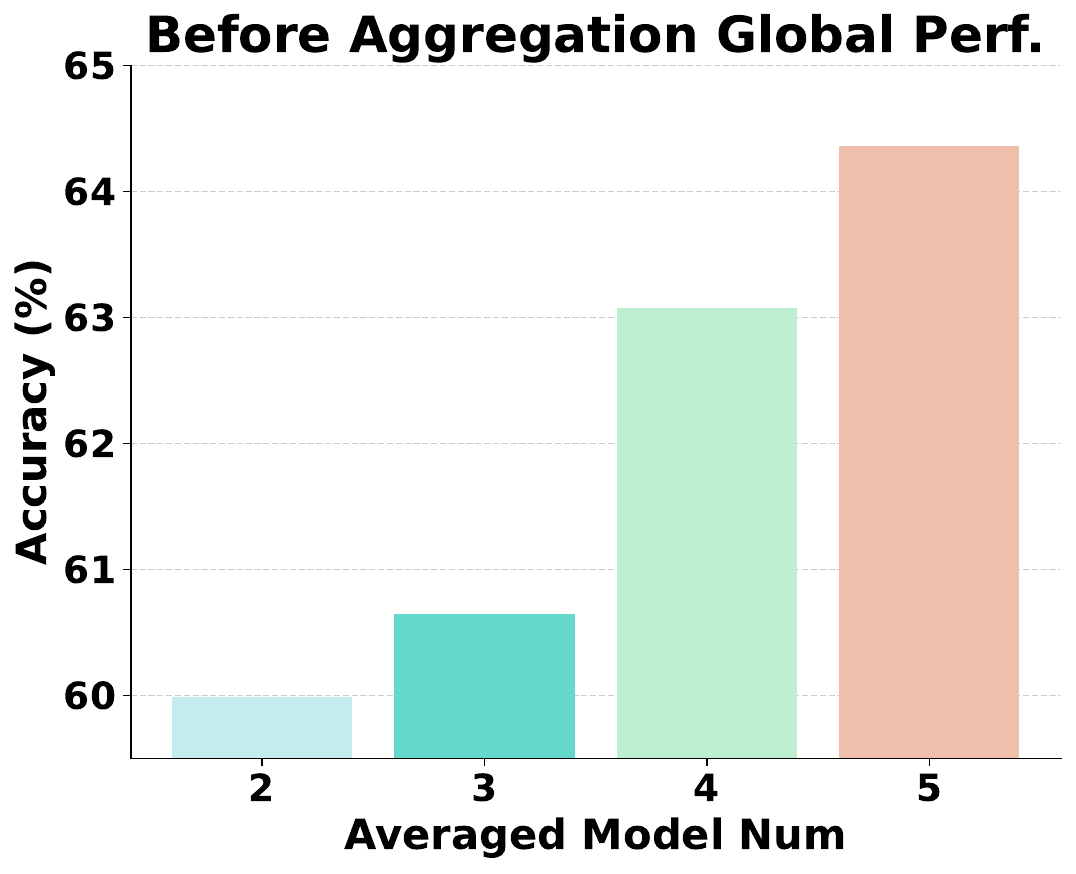}
    }
    \subfloat[] {
        \label{fig:sub:num_model_worst}
        \includegraphics[width=0.32\columnwidth]{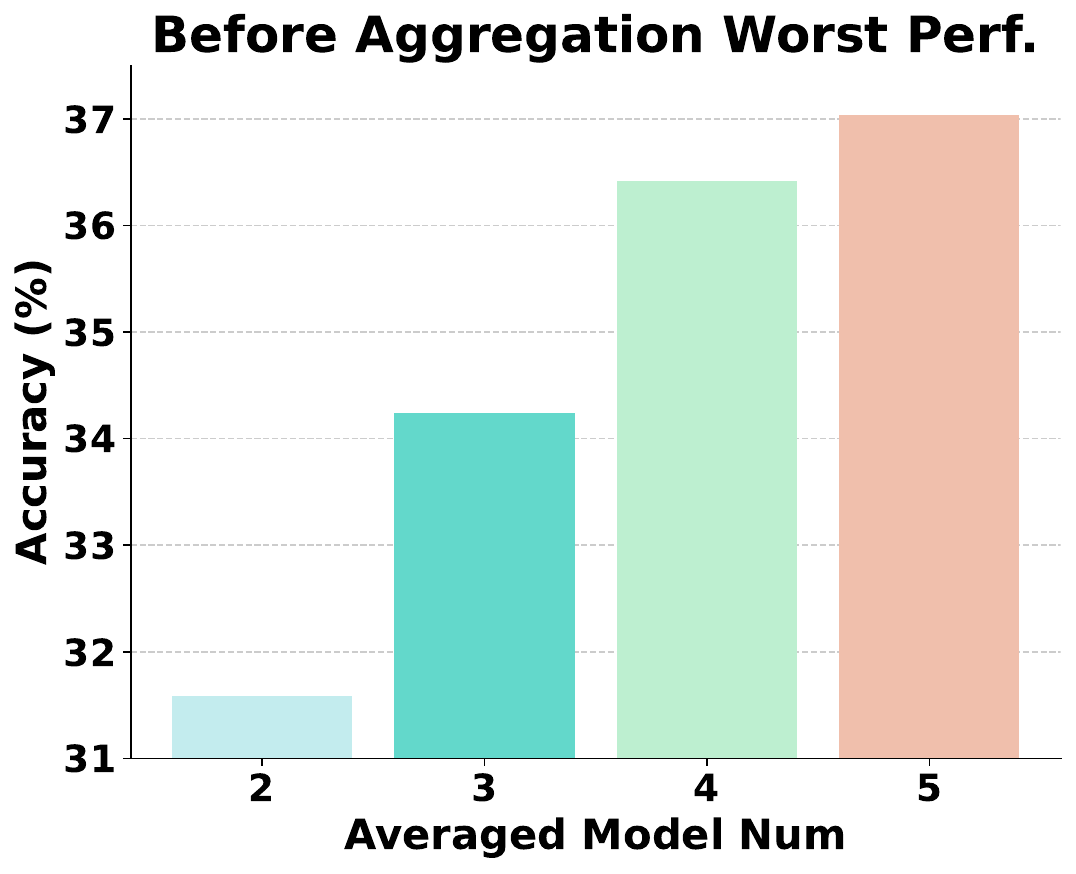}
    }
    \caption{\small Ablation studies on the impact of the number of averaged models on communication efficiency and performance variance.
    We evaluated the influence of varied model quantities on global and averaged local model performance, as well as generalization on the worst client.
    }
    \label{fig:num_model}
    \end{figure}

\noindent\textbf{Investigation on the number of averaged models.} To investigate the impact of the averaged model quantity on enhancing communication efficiency and cross-client generalization, we experiment with varied model quantities and evaluate their influence on global model performance, averaged local model performance\footnote{the average performance of local models of individual clients before aggregation on the overall client dataset}, and worst out-of-distribution (OOD) generalization performance on the other clients.
Fig.~\ref{fig:num_model} shows that increasing the number of averaged models can improve the model's OOD generalization ability and enhance the performance of the aggregated model. This similar upward trend supports our motivation for local model averaging.
We provide a more detailed analysis on connecting our proposed \ours{} and OOD generalization in appendix~\ref{sec:appendix_intuition}.
Additionally, we can observe that increasing the number of models in our method can improve both pre-aggregation and post-aggregation model performance.

\section{Conclusion}
\label{sec:conclusion}

\paragraph{Limitations and Broader Impact.}
Our method reduces communication rounds but trades off training memory and performance. Future work should explore more memory-efficient deployments. While mainly focused on vision tasks, extending to language and multimodal scenarios is promising.
Balancing performance and communication in healthcare FL is promising, but excessive reduction can impair critical medical decisions. Careful trade-off consideration is essential for reliable FL applications in sensitive areas.

\paragraph{Conclusion.} We propose an efficient method, Local Superior Soups (\ours{}), to reduce communication rounds in FL with pre-trained initialization, addressing the challenge of data heterogeneity.
By employing sequential model interpolation, connectivity preservation, and two regularization terms (diversity and affinity), the method allows for an increase in local training steps and a reduction in communication rounds while empirically maintaining strong aggregated performance.
This approach, tailored for pre-trained model adaptation in FL, offers training and inference efficiency, making it suitable for practical deployment in edge computing scenarios.
As the first step towards understanding and developing model soups-based methods in pre-trained models in FL, this study conducts experiments on benchmark datasets.
Our method attains superior performance with only a few rounds of communication and surpasses the performance of standard FL methods significantly across four datasets and under two distribution shift scenarios.

\section*{Acknowledgement}
M. Chen, Z. Wang and X. Li are grateful for the support of the Natural Science and Engineering Research Council of Canada (NSERC). M. Chen and X. Li are supported by the Canada CIFAR AI Chairs program, MITACS-CIFAR Catalyst Grant Program, NVIDIA Hardware Awards, the Digital Research Alliance of Canada, and Canada Foundation for Innovation (CFI). M. Jiang and Q. Dou are supported by the Research Grants Council of the Hong Kong Special Administrative Region (Project No. T45- 401/22-N).

\begingroup
\microtypesetup{expansion=false}
\bibliographystyle{plain}
\bibliography{neurips_2024}
\endgroup

\clearpage
\appendix
\textbf{Roadmap of Appendix.} We list the notation table in Section~\ref{sec:appendix_notation}.
We provide the assumptions and proofs of the convergence analysis in Section~\ref{sec:appendix_convergence}.
We present the theoretical intuition for the two regularization terms in Section~\ref{sec:appendix_intuition}.
We provide more related work in Section~\ref{sec:more_related_work} and further experiment details and results in Section~\ref{sec:appendix_expdetails}.

\section{Notation Table}
\label{sec:appendix_notation}

\begin{table}[htbp]
\caption{Important notations used in the paper.}
\label{tab:notation}
\centering
\small
\renewcommand{\arraystretch}{1.04}
\begin{tabular}{@{}p{0.25\linewidth}p{0.71\linewidth}@{}}
\toprule
Notation & Description \\
\midrule
\multicolumn{2}{@{}l}{\textit{Data and local training}} \\
$M,m,n$ & Number of clients; parameter dimension; samples per client. \\
$\mathcal X,\mathcal Y$ & Input and label spaces; $(x,y)$ is a labeled example, also denoted $\xi$ in the empirical loss. \\
$\mathcal D_i,\mathcal D$ & Client $i$'s training set; collection $\{\mathcal D_i\}_{i=1}^M$. \\
$\ell_i,\mathcal L_i,\mathcal L$ & Per-example loss; client empirical objective; global objective. \\
$p_i$ & Client aggregation weight, with $\sum_i p_i=1$. \\
$f,f_p,f_{p_i}$ & Model-parameter vector; shared round initialization; candidate parameters. \\
$\mathcal M,N,N_c$ & Local model pool; pooled or averaged member count; candidates added per round. \\
$f_{\rm interp},f_s,\alpha_i$ & Interpolated parameters; nonnegative coefficients satisfying $\sum_i\alpha_i=1$. \\
$\alpha$ (partition) & Dirichlet concentration used for label-shift data partitioning. \\
$\lambda_a,\lambda_d,\mathcal L_{\rm reg}$ & Affinity and diversity coefficients; regularized local objective. \\
$\operatorname{dist},\ell_{\rm affinity},\ell_{\rm diversity}$ & Euclidean parameter distance; affinity loss $\operatorname{dist}(f,f_p)$; diversity loss $\operatorname{dist}(f,\mathcal M)$, averaged over pool members. \\
$\eta,R,\tau$ & Learning rate; communication rounds; local update steps, per round in the local-SGD analysis and per candidate in Algorithm~\ref{alg:pseudo-code}. \\
\midrule
\multicolumn{2}{@{}l}{\textit{Convergence analysis}} \\
$f_i^{(r,k)},\overline f^{(r,k)}$ & Client parameters; arithmetic mean across clients at round $r$, step $k$. \\
$f^\star,d$ & Global minimizer; initialization distance from $f^\star$. \\
$\beta$ & Client loss-gradient Lipschitz constant in the convex analysis. \\
$\zeta$ & Uniform bound on client--global gradient disagreement. \\
$g_i^{(r,k)},\xi_i^{(r,k)},\sigma^2$ & Stochastic gradient; zero-mean gradient noise; upper bound on its conditional second moment. \\
$h_i^{(r,k)},c,q_i^{(r,k)}$ & Predictable gradient perturbation with $\|h_i^{(r,k)}\|_2\le c$; $q_i^{(r,k)}=\eta h_i^{(r,k)}$. \\
$a=\zeta+c$ & Combined client gradient-disagreement and perturbation bound. \\
$\mathcal F_{r,k},\mathbb E_k$ & History before the current gradients are drawn; conditional expectation given that history. \\
$b^{(r,k)}$ & Aggregate perturbation $M^{-1}\sum_i h_i^{(r,k)}$. \\
$\mathcal A_{\rm bias}$ & Signed average of $-\mathbb E\langle b^{(r,k)},\overline f^{(r,k+1)}-f^\star\rangle$ over the $\tau R$ updates. \\
$\epsilon$ & Expected average excess objective over post-update local-SGD iterates. \\
$K=M\tau R$ & Total stochastic-gradient evaluations in the local-SGD analysis. \\
\midrule
\multicolumn{2}{@{}l}{\textit{Weight-averaging analysis}} \\
$\theta_i,\bar\theta$ & Member weights; their arithmetic mean. \\
$P_T,(X,Y)$ & Target data distribution; an independent target input--label draw. \\
$h_\theta,Z_i,\bar Z$ & Prediction map; member prediction $Z_i=h_{\theta_i}$; mean prediction. \\
$V$ & $N^{-1}\sum_i\|\theta_i-\bar\theta\|^2$, the mean squared member spread. \\
$H$ & Common operator-norm Lipschitz constant of the prediction Jacobian on the member weights' convex hull. \\
$R_T(\theta),R_{\rm FE}$ & Squared target prediction risk; squared risk of the functional mean $\bar Z$. \\
$D_T$ & $\mathbb E[N^{-1}\sum_i\|Z_i(X)-\bar Z(X)\|^2]$, the predictive disagreement. \\
\bottomrule
\end{tabular}
\end{table}
\clearpage

\section{Convergence Analysis}
\label{sec:appendix_convergence}
\subsection{Formal Assumptions}
\label{app:theformal}

For positive integers $M,\tau,R$, all clients participate with weights
$p_i=1/M$ and a unit server
step, so $\mathcal L=M^{-1}\sum_i\mathcal L_i$.
Write $\overline f^{(r,k)}=M^{-1}\sum_i f_i^{(r,k)}$; synchronized rounds
satisfy $f_i^{(r,0)}=\overline f^{(r,0)}$ and
$\overline f^{(r+1,0)}=\overline f^{(r,\tau)}$.

\begin{assumption}[Convexity and smoothness]\label{assum:smooth}
Each $\mathcal L_i:\mathbb R^{\modelSize}\to\mathbb R$ is convex and
$\beta$-smooth, $\beta>0$. The global objective has a minimizer $f^\star$,
and $d=\|\overline f^{(0,0)}-f^\star\|_2$.
\end{assumption}

\begin{assumption}[Stochastic-gradient noise]\label{assum:bsg}
Let $\mathcal F_{r,k}$ contain the history before the current gradients
are drawn, and write
$g_i^{(r,k)}=\nabla\mathcal L_i(f_i^{(r,k)})+\xi_i^{(r,k)}$.
Conditionally on $\mathcal F_{r,k}$, the noises have zero mean and zero
cross-covariances across distinct clients, with
$\mathbb E[\|\xi_i^{(r,k)}\|_2^2\mid\mathcal F_{r,k}]\le\sigma^2$.
\end{assumption}

\begin{assumption}[Gradient dissimilarity]\label{assum:blgg}
For all $i,f$,
$\|\nabla\mathcal L_i(f)-\nabla\mathcal L(f)\|_2\le\zeta$.
\end{assumption}

\begin{assumption}[Perturbed update]\label{assum:breg}
The update is
$f_i^{(r,k+1)}=f_i^{(r,k)}-\eta(g_i^{(r,k)}+h_i^{(r,k)})$,
where $h_i^{(r,k)}$ is $\mathcal F_{r,k}$-measurable and
$\|h_i^{(r,k)}\|_2\le c$. Thus $c$ bounds a gradient-sized perturbation,
with $q_i^{(r,k)}=\eta h_i^{(r,k)}$.
\end{assumption}

Put $a=\zeta+c$ and $b^{(r,k)}=M^{-1}\sum_i h_i^{(r,k)}$. The signed
aggregate-bias remainder is
\begin{equation}\label{eq:aggregate_bias_remainder}
\mathcal A_{\rm bias}=-\frac1{\tau R}
\sum_{r=0}^{R-1}\sum_{k=0}^{\tau-1}
\mathbb E\langle b^{(r,k)},\overline f^{(r,k+1)}-f^\star\rangle.
\end{equation}
This term need not vanish.

\subsection{Proof of the Conditional Convergence Bound}

We follow the local-SGD argument of~\cite{DBLP:journals/corr/abs-2107-06917}, with
$0<\eta\le1/(4\beta)$ and $\mathbb E_k=\mathbb E[\,\cdot\mid\mathcal F_{r,k}]$.

\paragraph{Client drift.}
Within one round, suppress $r$ and write
\[
\begin{aligned}
\delta_k&=f_i^k-f_j^k,\qquad
v_k=\nabla\mathcal L(f_i^k)-\nabla\mathcal L(f_j^k),\\
e_i^k&=\nabla\mathcal L_i(f_i^k)-\nabla\mathcal L(f_i^k)+h_i^k,
\qquad\|e_i^k\|_2\le a.
\end{aligned}
\]
Cocoercivity gives $\langle v_k,\delta_k\rangle\ge\|v_k\|_2^2/\beta$.
Using
\[
\begin{aligned}
\|v_k+e_i^k-e_j^k\|_2^2&\le3\|v_k\|_2^2+6a^2,\\
4\eta a\|\delta_k\|_2&\le\|\delta_k\|_2^2/\tau+4\tau\eta^2a^2,
\end{aligned}
\]
since $3\eta^2-2\eta/\beta\le0$, the squared update and noise
assumptions give
\begin{equation}
\mathbb E_k\|\delta_{k+1}\|_2^2
\le(1+1/\tau)\|\delta_k\|_2^2
+10\tau\eta^2a^2+2\eta^2\sigma^2.
\end{equation}
Since $\delta_0=0$, the geometric sum for $k\le\tau$ is at most
$(e-1)\tau<1.8\tau$, yielding
\begin{equation}\label{eq:conditional_drift}
\mathbb E[\|f_i^{(r,k)}-f_j^{(r,k)}\|_2^2\mid\mathcal F_{r,0}]
\le18\tau^2\eta^2a^2+4\tau\eta^2\sigma^2.
\end{equation}
Jensen's inequality gives the same bound for each client's squared
deviation from $\overline f^{(r,k)}$.

\paragraph{Optimization progress.}
Write $y=\overline f^{(r,k)}$, $y^+=\overline f^{(r,k+1)}$,
$S=M^{-1}\sum_i\|f_i^{(r,k)}-y\|_2^2$, and
$\overline\xi=M^{-1}\sum_i\xi_i^{(r,k)}$.
The noise term satisfies
\[
\mathbb E_k\langle-\overline\xi,y^+-f^\star\rangle
=\eta\mathbb E_k\|\overline\xi\|_2^2\le\eta\sigma^2/M.
\]
Convexity, smoothness, and polarization of the averaged update give,
after discarding the nonpositive term
$(\beta-1/(2\eta))\mathbb E_k\|y^+-y\|_2^2$,
\begin{equation}\label{eq:conditional_progress}
\begin{split}
\mathbb E_k[\mathcal L(y^+)-\mathcal L(f^\star)]
\le&\frac{\|y-f^\star\|_2^2-\mathbb E_k\|y^+-f^\star\|_2^2}{2\eta}
+\frac{\eta\sigma^2}{M}+\beta S\\
&-\mathbb E_k\langle b^{(r,k)},y^+-f^\star\rangle.
\end{split}
\end{equation}
Summing over updates telescopes across synchronized rounds. Dropping the
final negative squared distance and using Eq.~\eqref{eq:conditional_drift} gives
\begin{equation}\label{eq:conditional_unoptimized}
\epsilon\le\frac{d^2}{2\eta\tau R}+\frac{\eta\sigma^2}{M}
+4\beta\tau\eta^2\sigma^2+18\beta\tau^2\eta^2a^2+\mathcal A_{\rm bias}.
\end{equation}
For the candidate learning rates $\eta_j$ in Theorem~\ref{thm:1}, omit
candidates with zero denominators. Bounding the initialization term by
$1/\min_j\eta_j\le\sum_j1/\eta_j$ and the other terms by their
corresponding candidates gives coefficients $2$, $1.5$, $4.5$, and
$18.5$. Rounding upward proves Eq.~\eqref{eq:main_thm:3}, with constants
$2$, $2$, $5$, and $19$.

\subsection{Proof of the Communication Proposition}
\label{app:proof}

Let $T_1,\ldots,T_4$ denote the four explicit terms in
Eq.~\eqref{eq:main_thm:3} after substituting $R=K/(M\tau)$, and put
$A=\sigma\sqrt K/(\beta dM^2)$. Their ratios are
\begin{equation}
\begin{aligned}
T_1/T_2&=1/(AM),\\
T_3/T_2&=\tfrac52(\tau/A)^{1/3},\\
T_4/T_2&=\tfrac{19}2\bigl(\tau/[(\sigma/a)\sqrt A]\bigr)^{2/3}
\quad(a>0).
\end{aligned}
\end{equation}
Under Proposition~\ref{th:prop}, $1\le\tau\le A$ and
$\tau\le(\sigma/a)\sqrt A$ bound these ratios by $1$, $5/2$, and $19/2$,
respectively. For $a=0$, omit the second restriction and $T_4=0$.
Thus $\epsilon\le28\sigma d/\sqrt K+\mathcal A_{\rm bias}$; the full
$O(\sigma d/\sqrt K)$ rate requires
$\mathcal A_{\rm bias}\le C\sigma d/\sqrt K$ with $C$ independent of $K$.

\section{Theoretical Intuitions.}
\label{sec:appendix_intuition}

\subsection{Decomposition of Generalization Bound}

Following \cite{DBLP:conf/nips/RameKRRGC22}, condition on $N\ge1$ learned
weights $\theta_i$ and draw $(X,Y)\sim P_T$ independently of the learning
process. Write $Z_i(x)=h_{\theta_i}(x)$,
$\bar Z=N^{-1}\sum_iZ_i$, $\bar\theta=N^{-1}\sum_i\theta_i$, and
$V=N^{-1}\sum_i\|\theta_i-\bar\theta\|^2$.
Define $R_T(\theta)=\mathbb E\|h_\theta(X)-Y\|^2$, finite for each member,
$R_{\rm FE}=\mathbb E\|\bar Z(X)-Y\|^2$, and
$D_T=\mathbb E[N^{-1}\sum_i\|Z_i(X)-\bar Z(X)\|^2]$.
Then
\begin{equation}
R_{\rm FE}=\frac1N\sum_iR_T(\theta_i)-D_T.
\label{eq:lss_member_risk_identity}
\end{equation}
If $Dh_\theta(x)$ is $H$-Lipschitz in operator norm on the weights'
convex hull, with a common $H$ for target-almost-every $x$, then
\begin{equation}
R_T(\bar\theta)\le\frac1N\sum_iR_T(\theta_i)-D_T
+HV\sqrt{R_{\rm FE}}+\frac{H^2V^2}{4}.
\label{eq:lss_diversity_locality_tradeoff}
\end{equation}
\begin{proof}
The identity $N^{-1}\sum_i\|Z_i-Y\|^2=\|\bar Z-Y\|^2
+N^{-1}\sum_i\|Z_i-\bar Z\|^2$ gives the risk decomposition.
Taylor expansion gives
\[
\begin{aligned}
h_{\theta_i}(x)&=h_{\bar\theta}(x)
+Dh_{\bar\theta}(x)(\theta_i-\bar\theta)+r_i(x),\\
\|r_i(x)\|&\le\tfrac H2\|\theta_i-\bar\theta\|^2.
\end{aligned}
\]
Averaging cancels the linear terms, so
$\|e(X)\|\le HV/2$ for $e=h_{\bar\theta}-\bar Z$.
Expanding $\|\bar Z-Y+e\|^2$ and using Cauchy--Schwarz gives
$R_T(\bar\theta)\le R_{\rm FE}+HV\sqrt{R_{\rm FE}}+H^2V^2/4$.
\end{proof}

Parameter distance is only a proxy for predictive disagreement $D_T$;
affinity encourages the locality $V$ used by this bound.

\begin{figure}[t]
    \centering
    \setlength{\abovecaptionskip}{-0.8mm}
    \includegraphics[width=1.0\columnwidth]{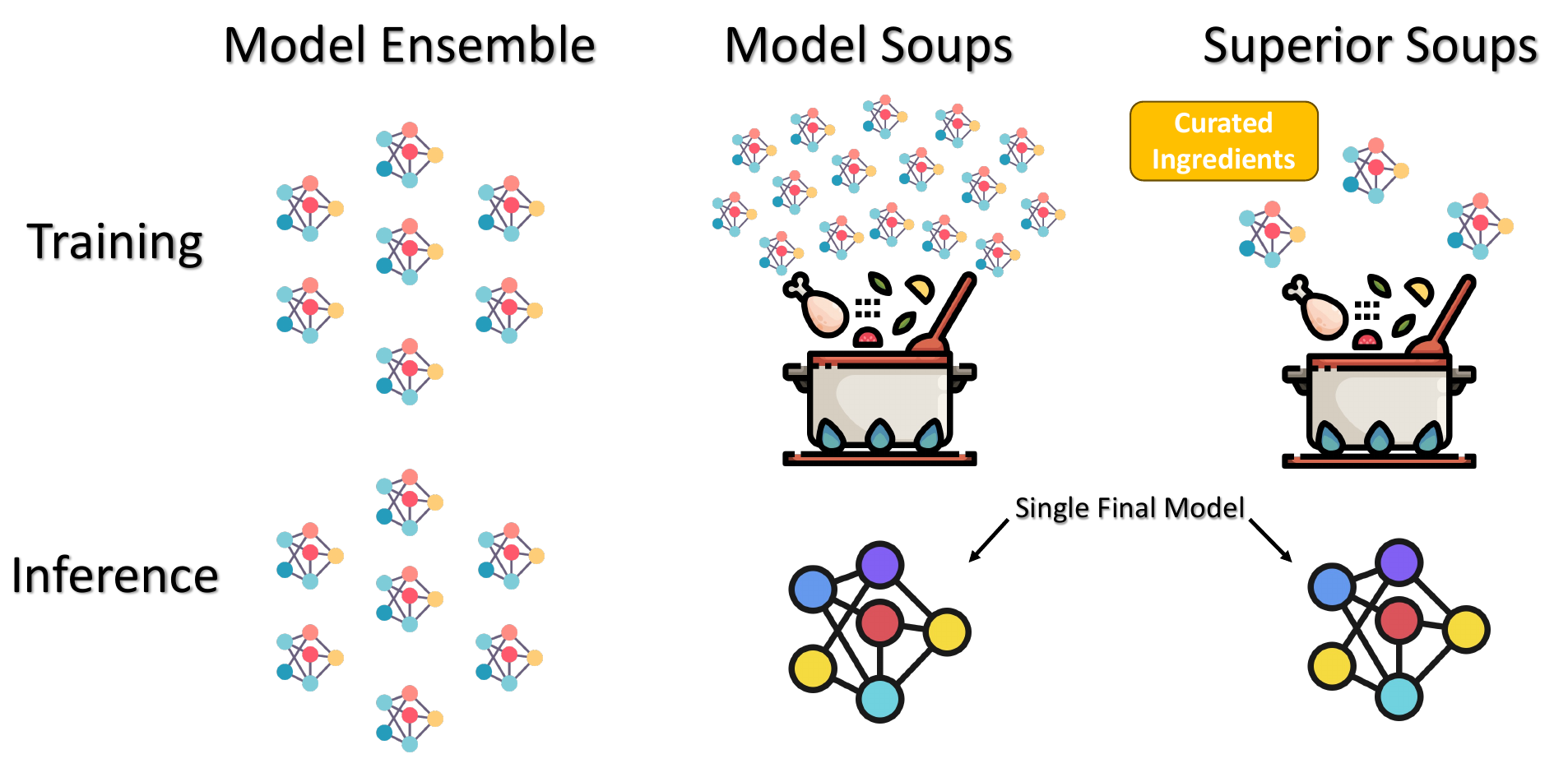}
    \caption{Comparison on model ensemble, model soups, and superior soups.}
    \label{fig:soup_comparison}
\end{figure}

\section{More Related Work}
\label{sec:more_related_work}

\subsection{Heterogeneous Federated Learning}
FL performance downgrading on Non-IID data is a critical challenge. A variety of FL algorithms have been proposed to handle this heterogeneous issue.
From an optimization perspective: FedProx~\cite{DBLP:conf/mlsys/LiSZSTS20} adds $L_2$ norm to the client model and the previous server model to regularize them. This helps to prevent the client models from diverging too far from the server model. Scaffold~\cite{DBLP:conf/icml/KarimireddyKMRS20} adds a variance reduction term to mitigate the ``clients-drift.''  MOON~\cite{DBLP:conf/cvpr/LiHS21} uses mode-level contrastive learning to stabilize local training by making the client models more robust to changes in the data distribution.
In addition, personalized FL~\cite{DBLP:journals/corr/abs-2103-00710} is another approach to achieving high local testing performance on Non-IID data. For aggregation perspective: FedBN~\cite{DBLP:conf/iclr/LiJZKD21} uses local batch normalization to alleviate the feature shift before averaging models.
For extreme communication efficient: In recent years, there have been some FL methods based on one-shot communication rounds. These methods typically use additional techniques on the server-side, such as using prediction ensembles \cite{DBLP:journals/corr/abs-1902-11175} instead of weight ensembles or generating data \cite{DBLP:conf/nips/0081C0L0DS022, heinbaugh2023datafree} from local models for centralized training, to improve the performance of the aggregated model. These methods are orthogonal to our client training-based approach. There are also works on few-round communication rounds in FL based on meta-learning frameworks \cite{DBLP:conf/nips/ParkHKSM21}, but the data partition used in the experimental setup may not be suitable for practical FL scenarios.

\subsection{Federated Fine-tuning and Model Interpolation}
\label{sec:related-ft-app}
Fine-tuning aims to achieve improved performance on the given task by leveraging the learned knowledge of the pre-trained model.
\cite{DBLP:journals/corr/abs-2211-00107} empirically study the impact of fine-tuning from a pre-trained model in FL and unsurprisingly find that starting from a pre-trained model reduces the training time required to reach a target error rate and enables the training of more accurate models than starting from random initialization.
\cite{DBLP:conf/cvpr/ZhangS0TD22} propose a knowledge distillation approach for fine-tuning the global model, called FedFTG.
In addition, fine-tuning in FL has been widely used in personalized FL to address Non-IID problems by having each user adapt the global model to personalized local models using their own data. For example, FedBABU~\cite{DBLP:journals/corr/abs-2106-06042} splits the model into body and head, then fine-tuning the head part for personalization.
\cite{DBLP:journals/corr/abs-2108-07313} propose FTFA and RTFA that start with a pre-trained model and then fine-tunes a small subset of model parameters using the FedAvg~\cite{DBLP:conf/aistats/McMahanMRHA17} algorithm.
However, this line of work focuses on optimizing local performance and ignores the generalization of global data. This can lead to a performance drop when we further update the global model from  the updated local models.
Weight averaging and model recycling are not only efficient ways to aggregate machine learning models but also present promising benefits of improving model generalizability.
Inspired by the linear mode connectivity property of neural networks trained with stochastic gradient descent (SGD)~\cite{DBLP:conf/nips/NagarajanK19, DBLP:conf/icml/FrankleD0C20}, Model Soups~\cite{DBLP:conf/icml/WortsmanIGRLMNF22} proposes to combine many independent runs with varied hyper-parameter configurations. Similarly, DiWA~\cite{DBLP:conf/nips/RameKRRGC22} utilizes this idea of Model Soups while theoretically analyzing the importance of training different models with diverse hyper-parameters within mild ranges. Soups-based methods~\cite{DBLP:conf/icml/WortsmanIGRLMNF22,DBLP:conf/nips/RameKRRGC22} rely on aggregating diverse models to improve model generalizability.
To induce greater diversity, some methods such as ~\cite{DBLP:conf/nips/MaddoxIGVW19} using a high constant learning rate, ~\cite{DBLP:conf/icml/WortsmanHGFR21} minimizing cosine similarity between weights, ~\cite{DBLP:conf/uai/IzmailovMKGVW19} using a tempered posterior and model Ratatouille~\cite{DBLP:journals/corr/abs-2212-10445} averages diverse model trained from auxiliary datasets.

\section{Experiment Details}
\label{sec:appendix_expdetails}

\subsection{Experimental Setup Details}

\noindent\textbf{Dataset.}
We validate the effectiveness of our proposed method with four datasets, FMNIST \cite{DBLP:journals/corr/abs-1708-07747}, CIFAR-10 \cite{krizhevsky2009learning}, Digit-5 \cite{DBLP:conf/icml/GaninL15, DBLP:conf/iclr/LiJZKD21}, and DomainNet \cite{DBLP:conf/iccv/PengBXHSW19}.
The Fashion-MNIST (FMNIST) dataset is a dataset of Zalando's article images consisting of a training set of $60,000$ examples and a test set of $10,000$ examples. Each example is a $28 \times 28$ grayscale image of a piece of clothing. The dataset is divided into $10$ classes: t-shirt/top, trouser, pullover, dress, coat, sandal, shirt, sneaker, bag, and ankle boot.
The CIFAR-10 dataset is a popular dataset for machine learning research. It consists of $60,000$ $32 \times 32$ color images divided into $10$ classes: airplane, automobile, bird, cat, deer, dog, frog, horse, ship, and truck. The dataset is split into $50,000$ training images and $10,000$ test images.
The Digit-5 dataset is a collection of five popular digit datasets, MNIST~\cite{DBLP:journals/spm/Deng12} (55000 samples), MNIST-M (55000 samples), Synthetic Digits~\cite{DBLP:conf/icml/GaninL15} (25000 samples), SVHN (73257 samples), and USPS (7438 samples). Each digit dataset includes a different style of 0-9 digit images.
The DomainNet dataset is a large-scale dataset of images collected from six different domains: clipart, infograph, painting, quickdraw, real, and sketch. The dataset contains $600,000$ images, each labeled with one of $345$ object categories. The images in the DomainNet dataset are of high quality and are diverse in terms of their content and style.

\noindent\textbf{Model.}
We used the pre-trained models from the timm repo \footnote[1]{https://github.com/huggingface/pytorch-image-models}, which are a collection of state-of-the-art deep learning models for computer vision tasks.
For our proposed \ours{}, we use Adam optimizer with a learning rate of $5\mathrm{e}{-4}$ , momentum $0.9$, and weight decay $5\mathrm{e}{-4}$.
The default number of averaged models is $4$.
Each model updates $8$ epoch then aggregates with the others.
The default affinity term coefficient is $3$ and diversity term coefficient is $3$.
We set the batch size to $64$ by default.
For vision transformer (ViT) \cite{DBLP:conf/iclr/DosovitskiyB0WZ21} model, we adopt ViT base model with $224 \times 224$ image size and $16 \times 16$ input patch size.
The ViT is a neural network architecture for image classification that uses a self-attention mechanism to learn the relationships between pixels in an image. ViT has been shown to achieve state-of-the-art results on a variety of image classification benchmarks, including ImageNet and CIFAR-10.

\noindent\textbf{Training Details.}
We implement all the methods in PyTorch, and we run all the experiments on an NVIDIA Tesla V100 GPU.
Unless otherwise specified, the model performance in the experiments below refers to the global model performance after aggregation on the server side.
For commonly used FL methods, due to the significant increase in local update steps that leads to worse convergence, we set their local update steps to $8$.

\noindent\textbf{Applying WAFT to FL Local Update.}
For SWA~\cite{DBLP:conf/uai/IzmailovPGVW18}, SWAD~\cite{DBLP:conf/nips/ChaCLCPLP21}, and our method \ours{}, we take more local update steps, with each model being averaged trained $8$ steps, and the default number of models to be averaged is $4$.
For the Model Soups~\cite{DBLP:conf/icml/WortsmanIGRLMNF22} method and DiWA~\cite{DBLP:conf/nips/RameKRRGC22}, we trained $32$ models and each model trained $8$ steps.
The hyper-parameter configuration for model selection includes learning rate ($[1\mathrm{e}{-4}, 5\mathrm{e}{-4}, 1\mathrm{e}{-5}]$), batch size ($[32, 64, 128]$), dropout rate ($[0.0, 0.1, 0.3]$), and weight decay $[5\mathrm{e}{-4}, 5\mathrm{e}{-5}, 5\mathrm{e}{-6}]$.
Each run randomly select one of the hyper-parameter options.
From each run of WAFT method, we take the weights of the epoch with maximum accuracy on the validation dataset, which follows the training distribution.

\subsection{Extended Experiment Results}
\label{sec:app_ex_exp}

\noindent\textbf{Arbitrarily increasing local steps cannot reduce communication rounds.}

From Table~\ref{table:app_local_step}, we can see that simply increasing local steps does not always lead to improved model performance. For FedAvg on the CIFAR10 dataset, increasing local steps beyond 8 actually results in a decrease in model performance.

\begin{table}[htpb]
    \centering
    \caption{FedAvg with different local steps: Label shift test accuracy after $R = 1$ communication rounds (CIFAR-10 with 5 Clients).
    }
    \label{table:app_local_step}
    \renewcommand\arraystretch{1.0}
    \resizebox{1.0\columnwidth}{!}{
    \begin{tabular}{l*8c}
        \toprule
        Method & Accuracy ($\tau = 1$) $\uparrow$  &  Accuracy ($\tau = 4$) $\uparrow$ &  Accuracy ($\tau = 8$) $\uparrow$ &  Accuracy ($\tau = 12$) $\uparrow$ & Accuracy ($\tau = 16$) $\uparrow$ \\
        \midrule
        FedAvg \cite{DBLP:conf/aistats/McMahanMRHA17} & $34.03 (2.84) $ & $49.08 (1.51) $ & $\textbf{58.34} (0.86) $ & $55.76 (0.82) $ & $53.21 (0.80) $ \\
        \bottomrule
    \end{tabular}}
\end{table}

\noindent\textbf{Computational and memory costs comparison.}

In Table~\ref{table:app_compute_mem_cost}, we report computational costs for the displayed configurations. This table does not include measured peak memory. The ``4x Mem.'' annotations in Tables~\ref{table:ls_main}--\ref{table:fs_main} refer to storage for four models. Since the computational overhead and memory usage of FedAvg and other used FL methods are nearly identical, we only present the data for FedAvg here. Similarly, as the computational overhead and memory usage for SWA and SWAD, as well as for Soups and DiWA, are also nearly the same, we only show the data for SWA and Soups methods.
Retaining several models can increase training memory, but a quantitative memory comparison requires measured values. The displayed configuration has a lower per-round time than the 32-model Soups adaptation and a higher per-round time than FedAvg. This is because other soups-based methods require training a large number of models repeatedly to achieve good model performance. For instance, Soups needs to train 32 models, whereas our method only requires training 4 models. If the number of models trained by Soups is reduced to just 4, it only brings about a 5\% improvement compared to FedAvg with a communication round of 1.

\begin{table}[htpb]
    \centering
    \caption{Computational costs of different methods (ResNet-18). Peak training memory is not reported in this table.
    }
    \label{table:app_compute_mem_cost}
    \renewcommand\arraystretch{1.0}
    \resizebox{1.0\columnwidth}{!}{
    \begin{tabular}{l*5c}
        \toprule
        Costs & FedAvg \cite{DBLP:conf/aistats/McMahanMRHA17} & SWA \cite{DBLP:conf/uai/IzmailovPGVW18} & Soups \cite{DBLP:conf/icml/WortsmanIGRLMNF22} & \textbf{\ours{} (2 models)} & \textbf{\ours{} (4 models)} \\
        \midrule
        MACs (G) & 1.82 & 1.82 & 1.82 & 2.73 & 4.55 \\
        \hline
        Train Time Per Epoch (s) & 2.66 & 2.73 & 2.66 & 12.27 & 20.43 \\
        Train Time Per Round (s) & 21.28 & 433.31 & 683.52 & 100.98 & 169.77 \\
        \bottomrule
    \end{tabular}}
\end{table}

\noindent\textbf{Robustness to input noise.}
In Table~\ref{table:app_smoothness}, we provide the performance degradation of trained models evaluating under varying levels of random noise. Smaller accuracy degradation under these input perturbations indicates greater robustness under this protocol; it does not estimate the parameter-gradient Lipschitz constant $\beta$. We can observe that our method exhibits greater robustness to noise perturbation.

\begin{table}[htpb]
    \centering
    \caption{Accuracy drop under random input noise at the displayed amplitudes, after $R=1$ and $R=3$ communication rounds on CIFAR-10.
    }
    \small
    \label{table:app_smoothness}
    \renewcommand\arraystretch{1.0}
    \resizebox{1.0\columnwidth}{!}{
    \begin{tabular}{l*5c}
        \toprule
        & \multicolumn{2}{c}{\textbf{CIFAR-10 ($4/255$)}} & \multicolumn{2}{c}{\textbf{CIFAR-10 ($8/255$)}} \\
        \cmidrule(lr){2-3} \cmidrule(lr){4-5}
        Method & Accuracy drop ($R = 1$) $\downarrow$  &  Accuracy drop ($R = 3$) $\downarrow$ &  Accuracy drop ($R = 1$) $\downarrow$ &  Accuracy drop ($R = 3$) $\downarrow$ \\
        \midrule
        FedAvg \cite{DBLP:conf/aistats/McMahanMRHA17} & $1.30$ & $ 1.17$ & $3.06$ & $2.93$ \\
        \textbf{\ours{}} & $0.89$ & $ 0.76$ & $2.37$ & $1.85$ \\
        \bottomrule
    \end{tabular}}
\end{table}

\noindent\textbf{\ours{} improves flatness of loss landscape.}
The sharpness measure utilized in the Table~\ref{table:app_hessian} computes the median of the dominant Hessian eigenvalue across all training set batches through the Power Iteration algorithm~\cite{DBLP:conf/bigdataconf/YaoGKM20}. This is a local empirical curvature proxy~\cite{DBLP:journals/corr/abs-2202-00661}, rather than the maximum curvature of the full objective or a uniform smoothness bound. For an indefinite Hessian, plain power iteration targets the eigenvalue of largest absolute magnitude; this estimate does not establish a uniform smoothness constant. As demonstrated in the presented table, our method has a smaller reported median curvature estimate than FedAvg under this measurement protocol. This comparison alone does not establish a generalization guarantee.
\begin{table}[htpb]
    \centering
    \caption{Median estimated minibatch Hessian curvature under the reported power-iteration protocol. This is a local empirical proxy.
    }
    \label{table:app_hessian}
    \renewcommand\arraystretch{1.0}
    \resizebox{1.0\columnwidth}{!}{
    \begin{tabular}{l*4c}
        \toprule
         & FedAvg $\downarrow$  & \ours{} (2 models) $\downarrow$ &  \ours{} (3 models) $\downarrow$ &  \ours{} (4 models) $\downarrow$ \\
        \midrule
        Hessian Eigenvalue & 193.18 & 147.20 & 136.67 & \textbf{119.14} \\
        \bottomrule
    \end{tabular}}
\end{table}

\noindent\textbf{Evaluation with more clients.}
To assess the effectiveness of our method in larger-scale client scenarios, we conducted an expanded experiment involving 50 clients.
From the Table~\ref{table:app_more_client}, we can observe that our proposed method maintains a significant advantage across different client scales, particularly when the number of communication rounds is small ($R=1$).

\begin{table}[htpb]
    \centering
    \caption{Different client numbers (5 Clients and 50 Clients): Label shift test accuracy after $R = 1$ and $R = 3$ communication rounds.
    }
    \label{table:app_more_client}
    \renewcommand\arraystretch{1.0}
    \resizebox{1.0\columnwidth}{!}{
    \begin{tabular}{l*8c}
        \toprule
        & \multicolumn{2}{c}{\textbf{CIFAR-10 (5 Clients)}} & \multicolumn{2}{c}{\textbf{CIFAR-10 (50 Clients)}} \\
        \cmidrule(lr){2-3} \cmidrule(lr){4-5}
        Method & Accuracy ($R = 1$) $\uparrow$  &  Accuracy ($R = 3$) $\uparrow$ &  Accuracy ($R = 1$) $\uparrow$ &  Accuracy ($R = 3$) $\uparrow$ \\
        \midrule
        FedAvg \cite{DBLP:conf/aistats/McMahanMRHA17} & $58.34 (0.86) $ & $66.74 (0.76) $ & $49.32 (0.93) $ & $68.39 (0.61)$ \\
        \hline
        \textbf{\ours{}}  & $\textbf{65.96} (1.50) $ & $\textbf{75.16} (1.07) $ & $\textbf{56.72} (0.53) $ & $\textbf{73.32} (0.46) $ \\
        \bottomrule
    \end{tabular}}
\end{table}

\begin{table}[htpb]
    \centering
    \caption{Different Network Architecture (ResNet-18 and ViT): Label shift test accuracy after $R = 1$ and $R = 3$ communication rounds.
    }
    \label{table:app_vit}
    \renewcommand\arraystretch{1.0}
    \resizebox{1.0\columnwidth}{!}{
    \begin{tabular}{l*8c}
        \toprule
        & \multicolumn{2}{c}{\textbf{CIFAR-10 (ResNet-18)}} & \multicolumn{2}{c}{\textbf{CIFAR-10 (ViT Base)}} \\
        \cmidrule(lr){2-3} \cmidrule(lr){4-5}
        Method & Accuracy ($R = 1$) $\uparrow$  &  Accuracy ($R = 3$) $\uparrow$ &  Accuracy ($R = 1$) $\uparrow$ &  Accuracy ($R = 3$) $\uparrow$ \\
        \midrule
        FedAvg \cite{DBLP:conf/aistats/McMahanMRHA17} & $58.34 (0.86) $ & $66.74 (0.76) $ & $60.35 (0.82) $ & $69.38 (0.51)$ \\
        \hline
        \textbf{\ours{}}  & $\textbf{65.96} (1.50) $ & $\textbf{75.16} (1.07) $ & $\textbf{67.48} (0.70) $ & $\textbf{76.81} (0.47) $ \\
        \bottomrule
    \end{tabular}}
\end{table}

\begin{table}[htpb]
    \centering
    \caption{Different Non-IID level (Dirichlet distribution $\alpha = 1.0$ and $\alpha = 0.1$): Label shift test accuracy after $R = 1$ and $R = 3$ communication rounds.
    }
    \label{table:app_alpha_level}
    \renewcommand\arraystretch{1.0}
    \resizebox{1.0\columnwidth}{!}{
    \begin{tabular}{l*8c}
        \toprule
        & \multicolumn{2}{c}{\textbf{CIFAR-10 ($\alpha = 1.0$)}} & \multicolumn{2}{c}{\textbf{CIFAR-10 ($\alpha = 0.1$)}} \\
        \cmidrule(lr){2-3} \cmidrule(lr){4-5}
        Method & Accuracy ($R = 1$) $\uparrow$  &  Accuracy ($R = 3$) $\uparrow$ &  Accuracy ($R = 1$) $\uparrow$ &  Accuracy ($R = 3$) $\uparrow$ \\
        \midrule
        FedAvg \cite{DBLP:conf/aistats/McMahanMRHA17} & $58.34 (0.86) $ & $66.74 (0.76) $ & $18.30 (2.25) $ & $45.85 (1.24)$ \\
        \hline
        \textbf{\ours{}}  & $\textbf{65.96} (1.50) $ & $\textbf{75.16} (1.07) $ & $\textbf{26.70} (1.62) $ & $\textbf{50.02} (0.82) $ \\
        \bottomrule
    \end{tabular}}
\end{table}

\begin{wrapfigure}{r}{0.45\textwidth}
    \centering
    \setlength{\abovecaptionskip}{0mm}
    \label{fig:fedaya}
    \includegraphics[width=0.45\columnwidth]{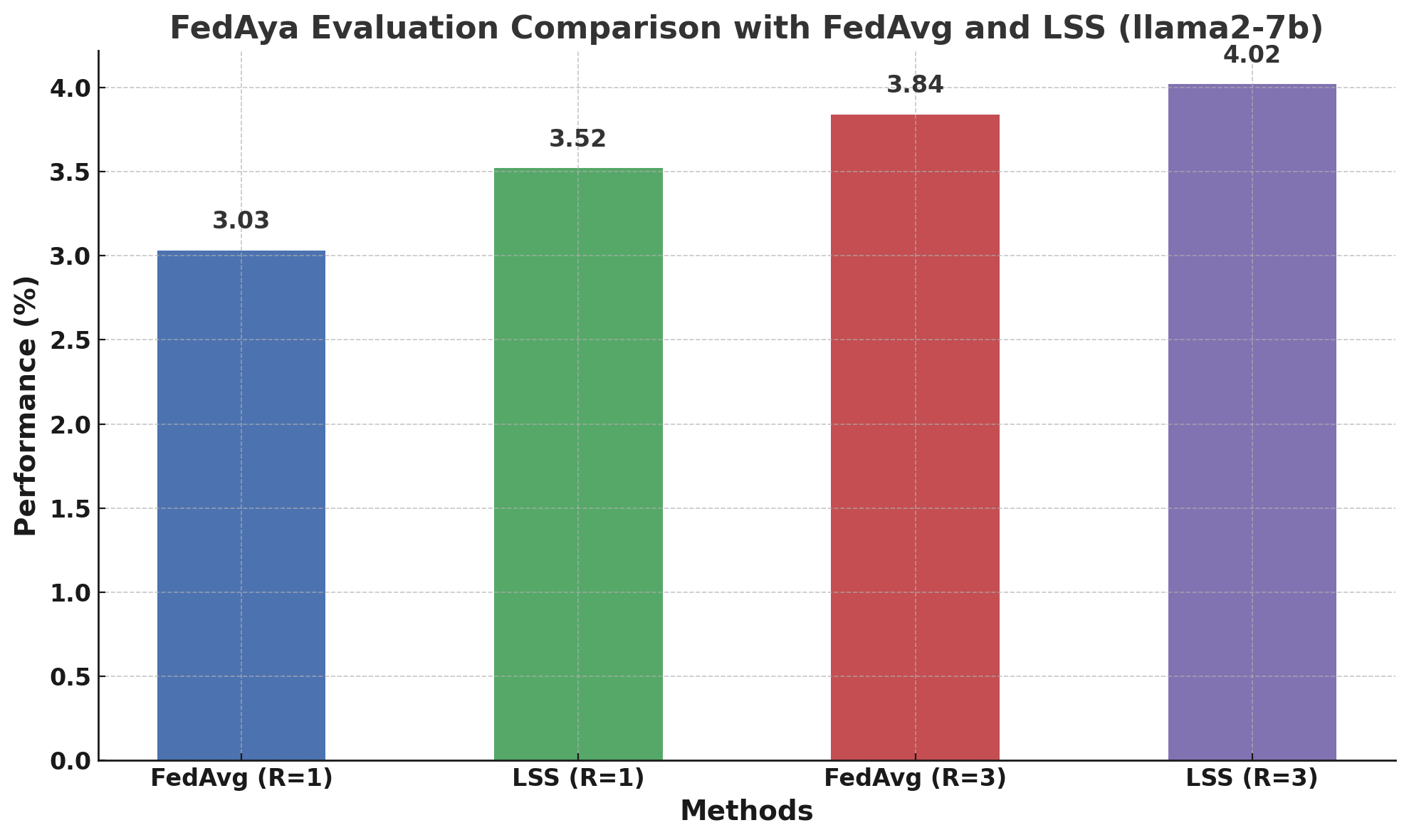}
    \caption{\small FedAya Evaluation Comparison with FedAvg and LSS. Our method, LSS, when applied to large language models for instruction tuning, achieves higher scores than the common FedAvg. This suggests that LSS is a promising approach for improving performance and convergence in federated learning settings for large language models, in addition to its success in image classification. Exploring the use of our method in a diverse set of complex LLM tasks is an interesting direction for future research.}
    \label{fig:fedaya_lss}
    \vspace{0mm}
\end{wrapfigure}

\noindent\textbf{Evaluation with ViT.}
To validate the effectiveness of our method across different network architectures, we conducted an expanded experiment using the Vision Transformer (ViT) model based on the Transformer architecture.
Upon observing the Table~\ref{table:app_vit}, it is evident that our method consistently enhances the communication efficiency of federated learning with ViT model architectures.

\noindent\textbf{Evaluation with different Non-IID level.}
To further comprehensively validate the effectiveness of our method under different levels of data heterogeneity, we conducted experiments on the CIFAR-10 dataset by adjusting the coefficients $\alpha$ of the Dirichlet distribution. We examined the performance of our method in scenarios with greater distribution variations.
Based on the Table~\ref{table:app_alpha_level}, it is evident that our method maintains a significant advantage in scenarios with larger data heterogeneity.

\noindent\textbf{Evaluation with different Initialized Models.}
To compare the performance of our method under different types of parameter initialization, we conducted experiments on the CIFAR-10 dataset using both pre-trained and random initialization. Table~\ref{table:app_comparison_init} shows that our method still maintains a significant advantage with random initialization, but it does not achieve the near-optimal performance seen with pre-trained initialization.

\begin{figure}[ht]
    \centering
    \setlength{\abovecaptionskip}{-0.8mm}
    \subfloat[] {
        \label{fig:sub:cifar_head_comm_fedprox}
        \includegraphics[width=0.22\columnwidth]{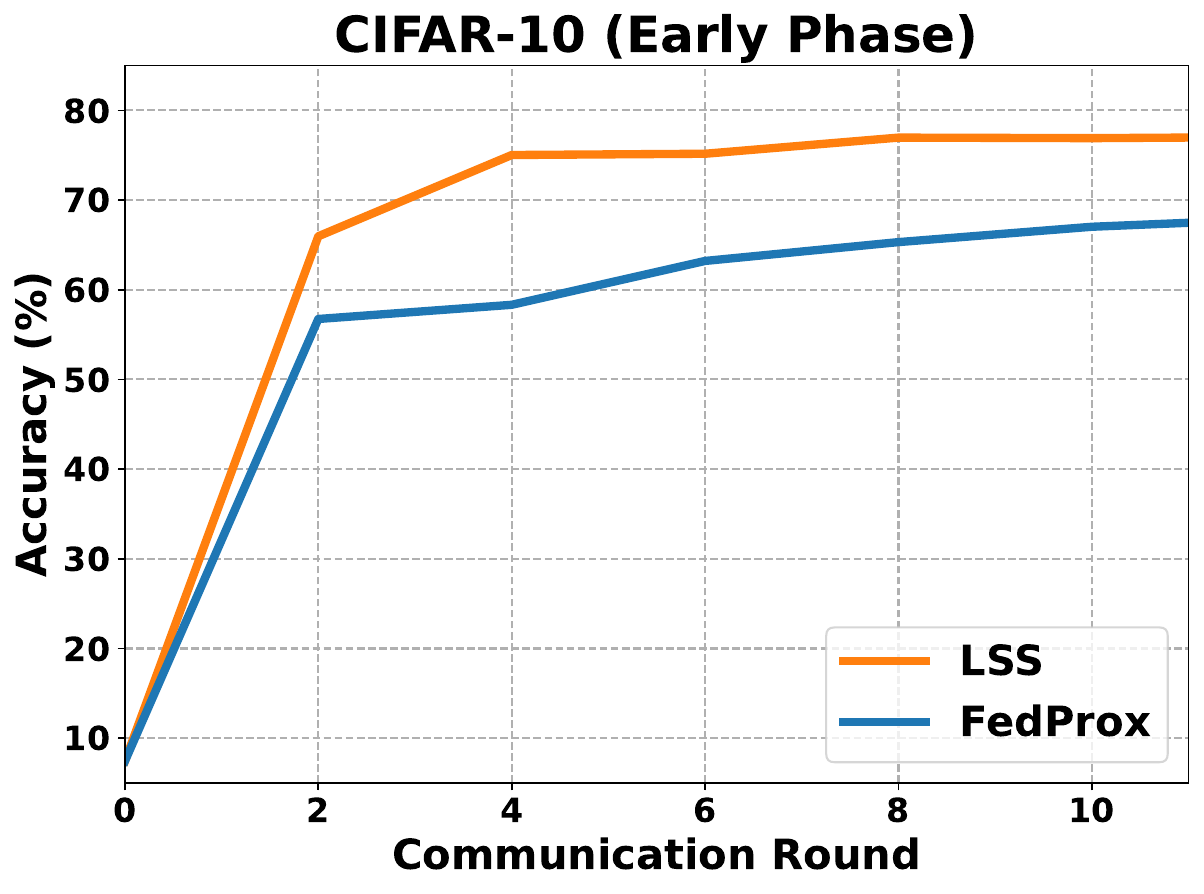}
    }
    \subfloat[] {
        \label{fig:sub:cifar_tail_comm_fedprox}
        \includegraphics[width=0.22\columnwidth]{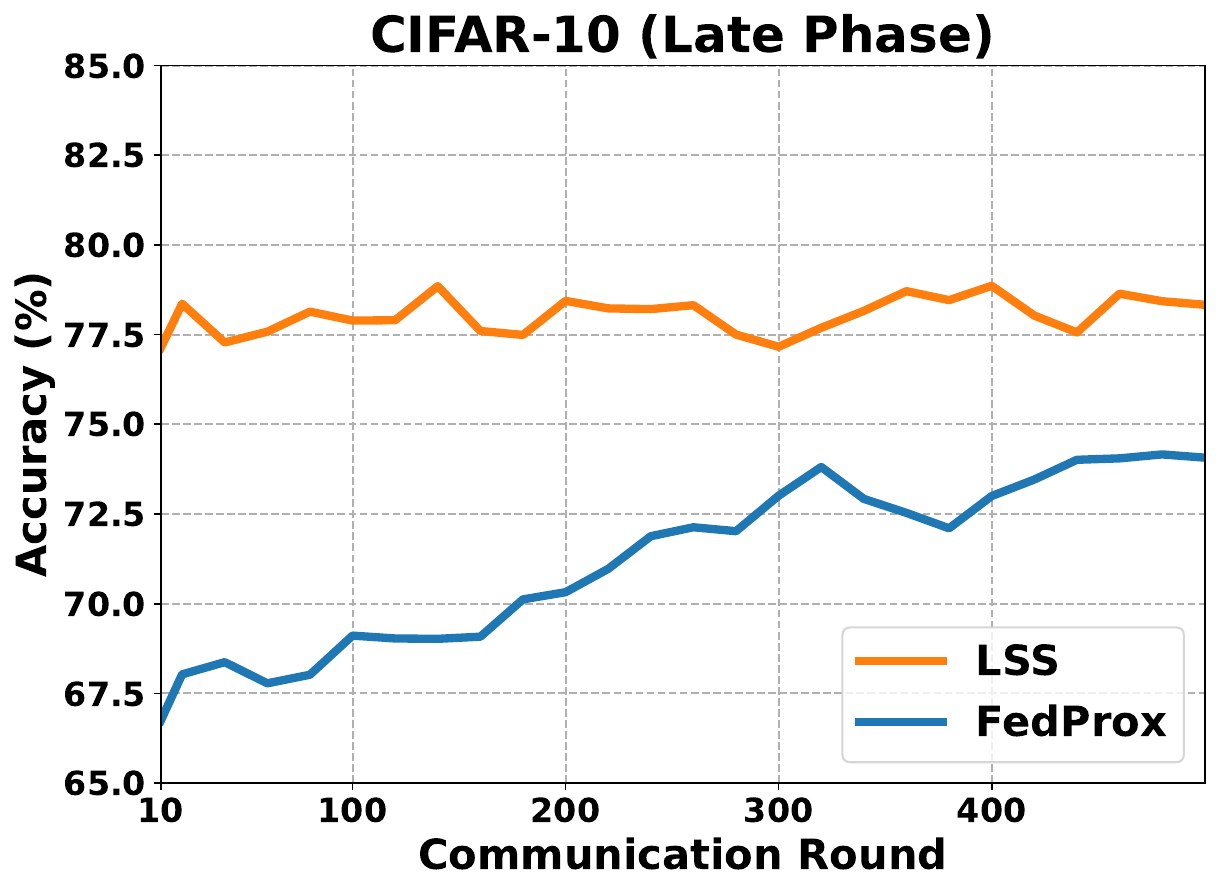}
    }
    \subfloat[] {
        \label{fig:sub:digit_head_comm_fedprox}
        \includegraphics[width=0.22\columnwidth]{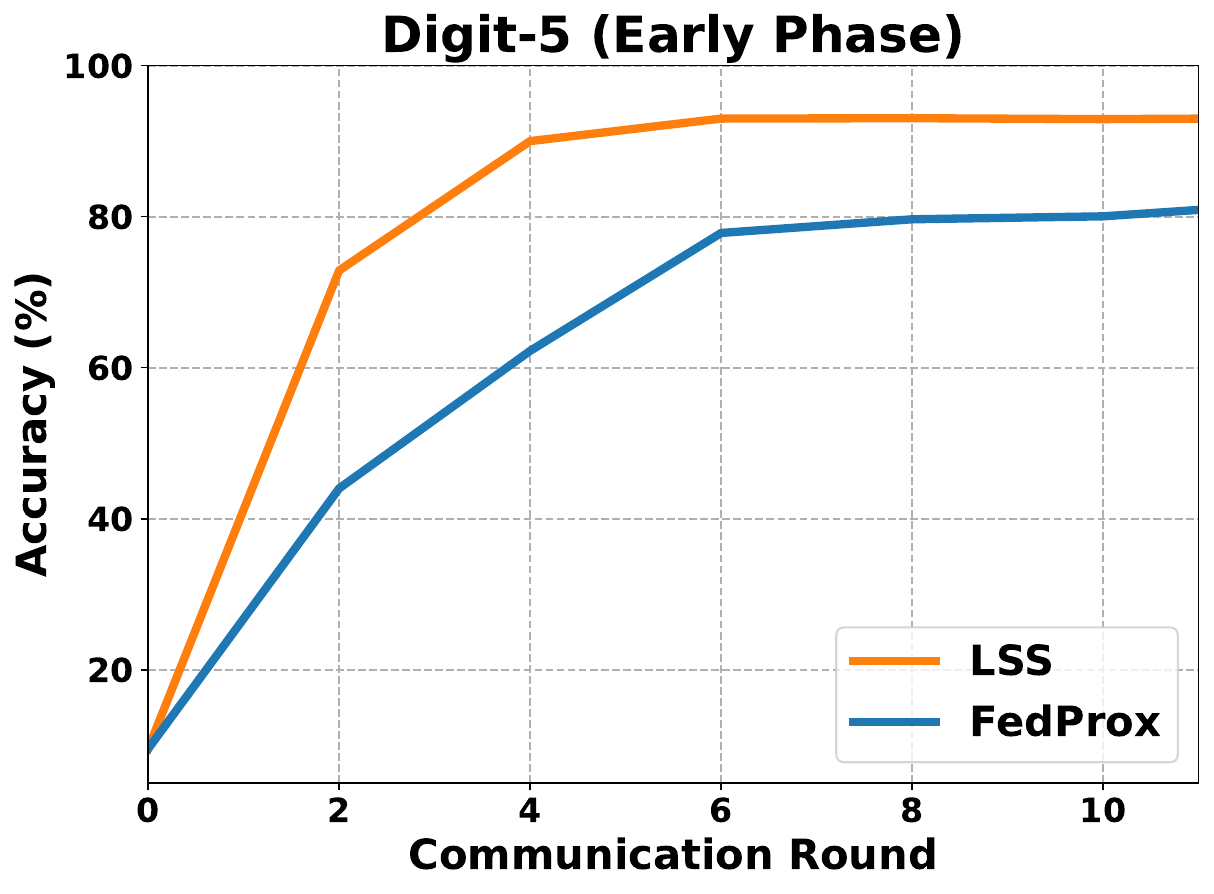}
    }
    \subfloat[] {
        \label{fig:sub:digit_tail_comm_fedprox}
        \includegraphics[width=0.22\columnwidth]{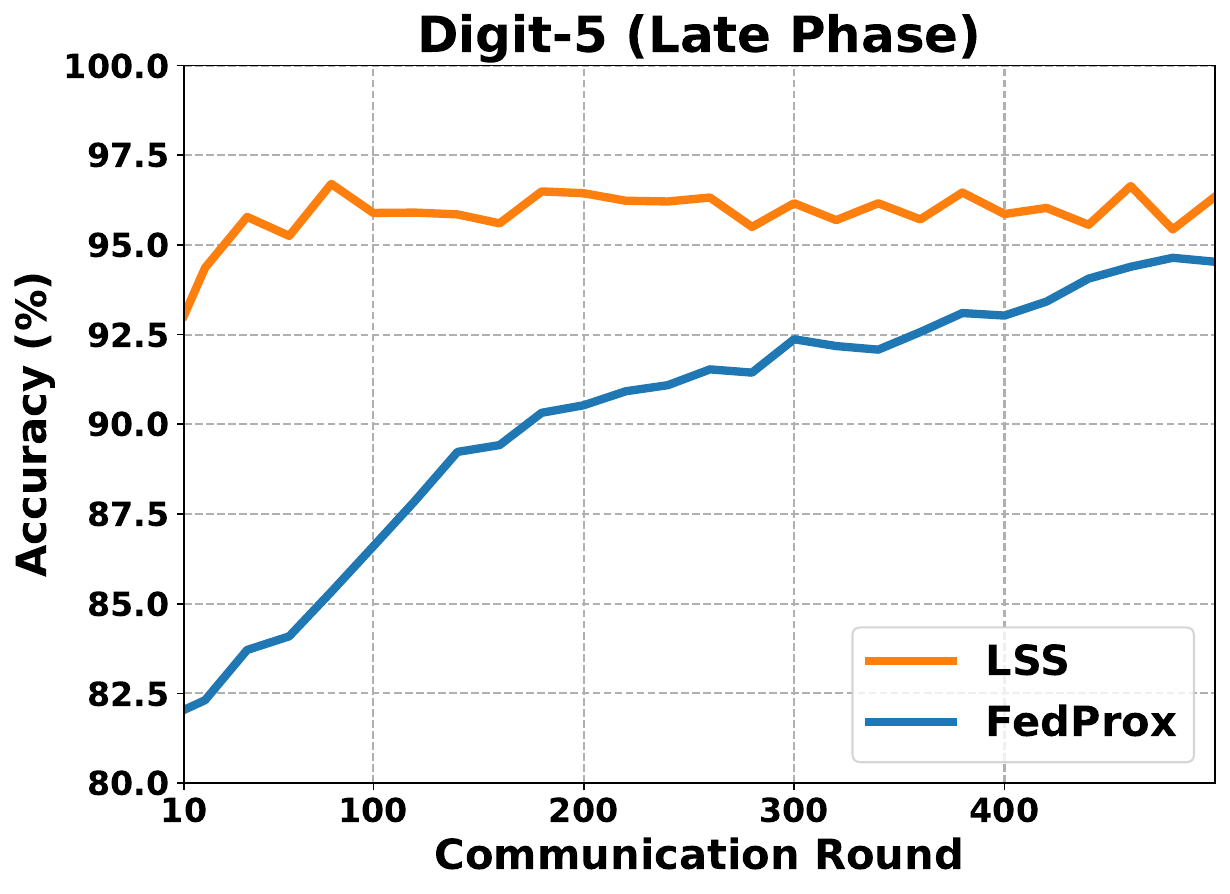}
    }
    \caption{\small Convergence comparison of our proposed \ours{} with FedProx. \ours{} also achieves high accuracy much earlier (around 6 to 8 rounds) than FedProx, which takes hundreds of communication rounds.}
    \label{fig:convergence_speed_fedprox}
    \vspace{-5mm}
\end{figure}

\noindent\textbf{Comparison of Convergence Speed Between FedProx and LSS.}
Fig.~\ref{fig:convergence_speed_fedprox} shows the accuracy of the testing during the early and late phases in terms of the number of communication rounds required to reach convergence. These results demonstrate that our method outperforms FedProx in both the early and late phases of federated learning.

\noindent\textbf{Evaluation of Large Language Models for Multilingual Instruction Tuning}

\textbf{Setup.}
We follow the setup of Fed-Aya~\cite{DBLP:journals/corr/abs-2406-04845}, which involves four iterative steps: server-to-client model downloading, local model training, client-to-server model uploading, and global model aggregation. For instruction tuning, we use the parameter-efficient fine-tuning technique, LoRA~\cite{DBLP:conf/iclr/HuSWALWWC22}, applied to the Llama-7b model.

\textbf{Dataset.}
We use the Aya dataset~\cite{DBLP:journals/corr/abs-2402-06619}, a multilingual instruction tuning dataset with annotations from contributors worldwide. Our experiments include 6 high-resource languages (English, Spanish, French, Russian, Portuguese, Chinese) and 2 low-resource languages (standard Arabic, Telugu). The dataset is filtered to include contributors with at least 100 annotations, resulting in 38 clients with a total of 25k data samples.

\textbf{Model.}
The model used for our experiments is the Llama2-7b~\cite{DBLP:journals/corr/abs-2307-09288}, fine-tuned using the LoRA technique. We evaluate the effectiveness of the training methods using an in-domain evaluation metric termed Ref-GPT4~\cite{DBLP:conf/nips/ZhengC00WZL0LXZ23}, where GPT-4o rates the generated responses against ground-truth responses. The score given by GPT-Ref ranges from 0 to 10. We adopt the same prompt template used in FedLLM-Bench~\cite{DBLP:journals/corr/abs-2406-04845}. The implementation of applying our method to LoRA is the same as that used in the ViT experiments (see Fig.~\ref{fig:vit_lora_digit5}) described earlier.

\textbf{Result.}
Our method, LSS, when applied to large language models for instruction tuning, achieves higher scores than the common FedAvg. This suggests that LSS is a promising approach for improving performance and convergence in federated learning settings for large language models, in addition to its success in image classification. Exploring the use of our method in a diverse set of complex LLM tasks is an interesting direction for future research.

\begin{table}[htpb]
    \centering
    \caption{Different model initialization (Pre-trained v.s. Random): Label shift test accuracy after $R = 1$ and $R = 3$ communication rounds. Result: It shows that our method still maintains a significant advantage with random initialization, but it does not achieve the near-optimal performance seen with pre-trained initialization.
    }
    \label{table:app_comparison_init}
    \renewcommand\arraystretch{1.0}
    \resizebox{1.0\columnwidth}{!}{
    \begin{tabular}{l*8c}
        \toprule
        & \multicolumn{2}{c}{\textbf{CIFAR-10 (Pre-trained)}} & \multicolumn{2}{c}{\textbf{CIFAR-10 (Random)}} \\
        \cmidrule(lr){2-3} \cmidrule(lr){4-5}
        Method & Accuracy ($R = 1$) $\uparrow$  &  Accuracy ($R = 3$) $\uparrow$ &  Accuracy ($R = 1$) $\uparrow$ &  Accuracy ($R = 3$) $\uparrow$ \\
        \midrule
        FedAvg \cite{DBLP:conf/aistats/McMahanMRHA17} & $58.34 (0.86) $ & $66.74 (0.76) $ & $14.83 (2.03) $ & $25.42 (0.71)$ \\
        \hline
        \textbf{\ours{}}  & $\textbf{65.96} (1.50) $ & $\textbf{75.16} (1.07) $ & $\textbf{30.64} (1.73) $ & $\textbf{37.86} (1.33) $ \\
        \bottomrule
    \end{tabular}}
\end{table}

\clearpage
\section*{NeurIPS Paper Checklist}

\begin{enumerate}

\item {\bf Claims}
    \item[] Question: Do the main claims made in the abstract and introduction accurately reflect the paper's contributions and scope?
    \item[] Answer: \answerYes{} %
    \item[] Justification: Yes, the main claims made in the abstract and introduction accurately reflect the paper’s contributions and scope. The concepts of federated learning settings are clearly introduced and defined in Sec.~\ref{sec:intro}.

\item {\bf Limitations}
    \item[] Question: Does the paper discuss the limitations of the work performed by the authors?
    \item[] Answer: \answerYes{}   %
    \item[] Justification: The paper acknowledges several limitations and areas for future research in Sec.~\ref{sec:conclusion}. Firstly, it highlights a trade-off between training memory and performance when reducing communication rounds in model merging. This indicates an awareness of the potential drawbacks of their method and suggests that further work is needed to make the method more training-memory efficient. Additionally, the paper notes that it focuses exclusively on vision-related tasks, suggesting that extending the findings to language tasks or multimodal scenarios would be a promising direction for future research.

\item {\bf Theory Assumptions and Proofs}
    \item[] Question: For each theoretical result, does the paper provide the full set of assumptions and a complete (and correct) proof?
    \item[] Answer: \answerYes{} %
    \item[] Justification: Yes, the paper meticulously outlines all relevant assumptions and presents comprehensive, logically sound proofs, ensuring the validity and reliability of each theoretical result. These elements are included in Sec.~\ref{sec:method} and Appendix.

    \item {\bf Experimental Result Reproducibility}
    \item[] Question: Does the paper fully disclose all the information needed to reproduce the main experimental results of the paper to the extent that it affects the main claims and/or conclusions of the paper (regardless of whether the code and data are provided or not)?
    \item[] Answer: \answerYes{} %
    \item[] Justification: The paper fully discloses all the information needed to reproduce the main experimental results. This information is provided in Sec.~\ref{sec:experiment}, with additional detailed information available in the appendix.

\item {\bf Open access to data and code}
    \item[] Question: Does the paper provide open access to the data and code, with sufficient instructions to faithfully reproduce the main experimental results, as described in supplemental material?
    \item[] Answer: \answerYes{} %
    \item[] Justification: Yes, the paper specifies all the training and test details necessary to understand the results, including data splits, hyperparameters, their selection process, and the type of optimizer used. This comprehensive detailing ensures that the experimental setup and outcomes are transparent and reproducible in Section~\ref{sec:experiment}.

\item {\bf Experimental Setting/Details}
    \item[] Question: Does the paper specify all the training and test details (e.g., data splits, hyperparameters, how they were chosen, type of optimizer, etc.) necessary to understand the results?
    \item[] Answer: \answerYes{} %
    \item[] Justification: The paper specifies all the training and test details necessary to understand the results, including data splits, hyperparameters, their selection process, and the type of optimizer used. These details are comprehensively documented in Sec.~\ref{sec:experiment} to ensure clarity and reproducibility.

\item {\bf Experiment Statistical Significance}
    \item[] Question: Does the paper report error bars suitably and correctly defined or other appropriate information about the statistical significance of the experiments?
    \item[] Answer: \answerYes{} %
    \item[] Justification: Yes, the paper reports error bars and other appropriate information about the statistical significance of the experiments in Sec.~\ref{sec:experiment}. These details are suitably and correctly defined to ensure the reliability and validity of the reported results.

\item {\bf Experiments Compute Resources}
    \item[] Question: For each experiment, does the paper provide sufficient information on the computer resources (type of compute workers, memory, time of execution) needed to reproduce the experiments?
    \item[] Answer: \answerYes{} %
    \item[] Justification: Yes, the paper provides sufficient information on the computer resources needed to reproduce the experiments in Sec.~\ref{sec:experiment}. This includes details on the type of compute workers, memory requirements, and execution time, ensuring that others can accurately replicate the experimental setup and results.

\item {\bf Code Of Ethics}
    \item[] Question: Does the research conducted in the paper conform, in every respect, with the NeurIPS Code of Ethics \url{https://neurips.cc/public/EthicsGuidelines}?
    \item[] Answer: \answerYes{} %
    \item[] Justification: Yes, the research conducted in the paper conforms, in every respect, with the NeurIPS Code of Ethics. The paper adheres to the guidelines on responsible release and publication strategy, ensuring that all necessary safeguards are in place for controlled use of the model (see Sec.~\ref{sec:experiment}).

\item {\bf Broader Impacts}
    \item[] Question: Does the paper discuss both potential positive societal impacts and negative societal impacts of the work performed?
    \item[] Answer: \answerYes{} %
    \item[] Justification: es, the paper discusses both potential positive and negative societal impacts of the work performed. The discussion includes an analysis of how the research can benefit society and also addresses possible adverse effects, ensuring a balanced and comprehensive evaluation of the societal implications in Sec.~\ref{sec:conclusion}.

\item {\bf Safeguards}
    \item[] Question: Does the paper describe safeguards that have been put in place for responsible release of data or models that have a high risk for misuse (e.g., pretrained language models, image generators, or scraped datasets)?
    \item[] Answer: \answerNA{} %
    \item[] Justification: The paper does not describe safeguards for the responsible release of data or models that have a high risk for misuse, such as pretrained language models, image generators, or scraped datasets. Because we does not focus the release of such assets, we focus model training techniques. (see Sec.~\ref{sec:experiment}).

\item {\bf Licenses for existing assets}
    \item[] Question: Are the creators or original owners of assets (e.g., code, data, models), used in the paper, properly credited and are the license and terms of use explicitly mentioned and properly respected?
    \item[] Answer: \answerNA{} %
    \item[] Justification: The paper does not utilize external assets such as code, data, or models from other creators or original owners. Therefore, there are no specific credits, licenses, or terms of use that need to be mentioned or respected.

\item {\bf New Assets}
    \item[] Question: Are new assets introduced in the paper well documented and is the documentation provided alongside the assets?
    \item[] Answer: \answerNA{} %
    \item[] Justification: The paper does not introduce any new assets such as code, data, or models. Therefore, there is no documentation provided alongside new assets.

\item {\bf Crowdsourcing and Research with Human Subjects}
    \item[] Question: For crowdsourcing experiments and research with human subjects, does the paper include the full text of instructions given to participants and screenshots, if applicable, as well as details about compensation (if any)?
    \item[] Answer: \answerNA{} %
    \item[] Justification: The paper does not involve crowdsourcing experiments or research with human subjects. Therefore, it does not include instructions given to participants, screenshots, or details about compensation.

\item {\bf Institutional Review Board (IRB) Approvals or Equivalent for Research with Human Subjects}
    \item[] Question: Does the paper describe potential risks incurred by study participants, whether such risks were disclosed to the subjects, and whether Institutional Review Board (IRB) approvals (or an equivalent approval/review based on the requirements of your country or institution) were obtained?
    \item[] Answer: \answerNA{} %
    \item[] Justification: The paper does not involve any study participants, crowdsourcing experiments, or research with human subjects. Therefore, it does not describe potential risks incurred by study participants, disclose such risks to subjects, or obtain Institutional Review Board (IRB) approvals.

\end{enumerate}

\end{document}